\documentclass[sn-apa]{sn-jnl}

\usepackage[utf8]{inputenc} 
\usepackage[T1]{fontenc}    
\usepackage{hyperref}       
\usepackage{url}            
\usepackage{booktabs}       
\usepackage{nicefrac}       
\usepackage{microtype}      
\usepackage{lipsum}		
\usepackage{graphicx}
\usepackage{natbib}
\usepackage{doi}
\usepackage{bbm}

\theoremstyle{thmstyleone}%
\newtheorem{theorem}{Theorem}
\newtheorem{definition}{Definition}

\newtheorem{lemma}{Lemma}

\newtheorem{example}{Example}
\usepackage{etoolbox}

\algnewcommand{\LineComment}[1]{\State $\triangleright$ #1}

\usepackage{caption}
\usepackage{subfigure}
\DeclareMathOperator*{\argmax}{arg\,max}
\DeclareMathOperator*{\argmin}{arg\,min}

\usepackage{ifthen}
\usepackage{bm}
\usepackage{amsfonts}
\usepackage{stackengine}
\usepackage{cleveref}
\let\oldhref\href
\renewcommand{\href}[2]{\oldhref{#1}{\hbox{#2}}}

\usepackage[british]{babel}
\usepackage{hyphenat}

\jyear{2021}

\begin{document}
\title[Explicit Explore, Exploit, or Escape ($E^4$)]{Explicit Explore, Exploit, or Escape ($E^4$): near-optimal safety-constrained reinforcement learning in polynomial time}
\author*[1]{\fnm{David M.} \sur{Bossens}}\email{d.m.bossens@soton.ac.uk}

\author[1]{\fnm{Nicholas} \sur{Bishop}}\email{nb8g13@soton.ac.uk}

\affil*[1]{\orgdiv{School of Electronics and Computer Science}, \orgname{University of Southampton}, \postcode{SO17 1BJ}, \country{United Kingdom}}

\abstract{
In reinforcement learning (RL), an agent must explore an initially unknown environment in order to learn a desired behaviour.
When RL agents are deployed in real world environments, safety is of primary concern. Constrained Markov decision processes (CMDPs) can provide long-term safety constraints; however, the agent may violate the constraints in an effort to explore its environment. 
This paper proposes a model-based RL algorithm called Explicit Explore, Exploit, or Escape ($E^{4}$), which extends the Explicit Explore or Exploit ($E^{3}$) algorithm to a robust CMDP setting. $E^4$ explicitly separates exploitation, exploration, and escape CMDPs, allowing targeted policies for policy improvement across known states, discovery of unknown states, as well as safe return to known states. $E^4$ robustly optimises these policies on the worst-case CMDP from a set of CMDP models consistent with the empirical observations of the deployment environment. 
Theoretical results show that $E^4$ finds a near-optimal constraint-satisfying policy in polynomial time whilst satisfying safety constraints throughout the learning process. We then discuss $E^4$ as a practical algorithmic framework, including robust-constrained offline optimisation algorithms, the design of uncertainty sets for the transition dynamics of unknown states, and how to further leverage empirical observations and prior knowledge to relax some of the worst-case assumptions underlying the theory.}

\maketitle

\footnote{Cite as: Bossens, D.M., Bishop, N. Explicit Explore, Exploit, or Escape ($E^4$): near-optimal safety-constrained reinforcement learning in polynomial time. Machine Learning (2022). https://doi.org/10.1007/s10994-022-06201-z}

\section{Introduction}
As machine learning methods are increasingly deployed in real-world applications, their safety is of foremost importance. 
Reinforcement learning (RL) is a particularly relevant example since in RL, an agent explores an initially unknown environment through observation, action, and reward.
Among the various challenges to safe artificial intelligence (see \cite{Everitt2018,Amodei2016}), safe exploration and robustness to mismatches between training and application environment are of particular importance when applying RL algorithms in safety-critical missions. Unsafe exploration may occur when an RL system explores a new region of the state space, and in doing so performs actions with disastrous consequences. A mismatch between training and application environment may happen when the model within a model-based RL system is incorrect on some states, which can result in the system learning behaviours that are safe in simulation but dangerous in the real world. More generally, desirable behaviour in a complex world comes with various \textit{safety constraints} such as avoiding damage or long-term wear-and-tear or following legal and social norms to avoid harm.

RL research has traditionally been studied from two angles, namely model-free and model-based RL, within the framework of Markov decision processes (MDPs). In model-free RL (e.g. \cite{Watkins1992,Rummery1994}), the RL agent has no model of the environment and only learns the long-term cumulative reward associated with the action taken in a given state. Deep model-free RL (e.g. \cite{Mnih2015,Schulman2017a}) additionally incorporates the expressive power of deep neural networks for high performance across a variety of simulation environments. Such methods do not consider safety throughout the exploration process, and therefore are likely to violate safety criteria. In model-based RL (e.g. \cite{Kearns2002,Brafman2002,Strehl2006}), the RL agent learns a model of the environment, including the transitions between states and the reward function, which can then be used to compute the long-term cumulative reward. Since learning in the real world may come with high-cost failures, a key benefit of model-based RL is the use of offline optimisation, which can sample trajectories from the environment model  without requiring too many real world samples and therefore, failures. Consequently, model-based RL provides improved sample complexity and may also be beneficial for safety. Before model-based RL can be used in long-term safety-critical applications, at least four requirements must be satisfied, namely constraint-satisfaction, safe exploration,  robustness to model errors, and applicability in non-episodic environments.

Constraint-satisfaction means that the agent must not violate any of the safety constraints defined by the user. The main framework for dealing with such constraints in this general manner is constrained Markov decision processes (CMDPs) \citep{Altman1999}, which aims to optimise the long-term  reward subject to constraints over a long-term constraint-cost function; in particular, the RL agent is given a maximum budget of cumulative constraint-cost which cannot be exceeded. Traditionally, CMDPs are solved by linear programming, and more recently by policy optimisation methods for high-dimensional control problems  \citep{Achiam2017}. 

Safe exploration means that the agent must not venture too far into unknown or dangerous states and ensure a timely return to safe known states. Within CMDPs, there are a variety of approaches to safe return based on a few additional assumptions. For control problems where the aim is to stay stable in a particular goal state, Lyapunov functions can be used with additional assumption of continuity on the policy and environment dynamics \citep{Chow2018,Berkenkamp2017}. An approach for general CMDPs is to define a supervising agent which can return the agent safely \citep{Turchetta2020}. However, this defers the problem of safe exploration to the supervising agent, which needs to be trained on realistic data. A model-free approach for safe exploration involves imposing an entropy constraint on policies to ensure exploration \citep{Yang2021}. As an alternative to the CMDP framework, one can also consider safety in terms of ensuring performance improves on a known policy \citep{Thomas2015,Garcelon2020}; in this case safe exploration can be ensured using off-policy evaluation \citep{Thomas2015}, which samples only from a known baseline policy, or by ensuring the online exploration is monotonically improving the performance \citep{Garcelon2020}.

Robustness to model errors means that the agent must be prepared for the worst-case when its model is inaccurate. This approach has been considered primarily by the framework of robust Markov decision processes, which accounts for uncertainty on the transition dynamics model \citep{Iyengar2005,Nilim2005,Wiesemann2013} and which has been integrated in CMDPs by \cite{Russel2020}. Others additionally incorporate uncertainty over the reward model with applications to non-stationary MDPs \citep{Jaksch2010,Lecarpentier2019a}, although these approaches are in the unconstrained setting and the optimistic bias in \cite{Jaksch2010} is likely to be unsafe. A setting with both reward and constraint-cost uncertainty has also been considered but this assumes known transition dynamics \citep{Zheng2020}.

Applicability in non-episodic (also called ``continuing'') environments means that the agent will be embedded in a long-term environment with no resets. The potential of continued sequential dependencies makes non-episodic environments challenging; it is therefore no surprise that non-episodic environments became a topic for recent benchmark formulations (see \cite{Naik2021,Platanios2020}). In safety-critical settings, the continued lifetime means that any failure is unforgiving-- unlike in the episodic video games. Therefore, the above-mentioned challenge of returnability becomes critically important in such environments. Somewhat similar to \cite{Turchetta2020}, \cite{Eysenbach2018}  defines a separate policy for resetting, with the main aim to abort and return to an initial state, mimicking the episodic setting without manual resets being required. Similar principles may be applied to the safety-constrained setting as well, where one could potentially target return to  a larger set of safe and known states.

With respect to the above requirements, Explicit Explore or Exploit ($E^3$) \citep{Kearns2002}, a model-based RL algorithm for near-optimal RL with polynomial sample and time complexity guarantees, provides a unique starting point for safe model-based RL. $E^3$ approximates the MDP with which the agent is interacting and the algorithm accounts for model errors and the resulting value estimate errors by distinguishing between known states, which haven been estimated correctly, and unknown states, which have not been estimated correctly. $E^3$ also provides a natural way to deal with exploration and exploitation in a non-episodic environment, namely by alternating limited-step trajectories, each of which is long enough to assess value function statistics correctly, and then explicitly choosing either an exploration policy or an exploitation policy. The exploitation policy is chosen when an optimal policy is available from the given starting state of the trajectory, and an exploration policy is chosen otherwise, in an attempt to find an unknown state. Recent work \citep{Henaff2019} has also shown the practical feasibility of the approach for continuous state spaces, comparing favourably to state-of-the-art deep RL.

\begin{figure}
\includegraphics[width=0.13\textwidth]{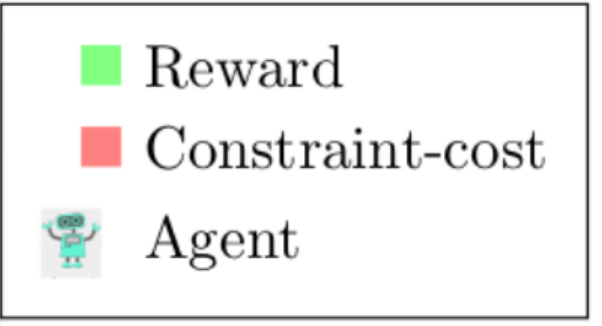}  \subfigure[Exploit (Start)]{\includegraphics[height=0.13\textheight]{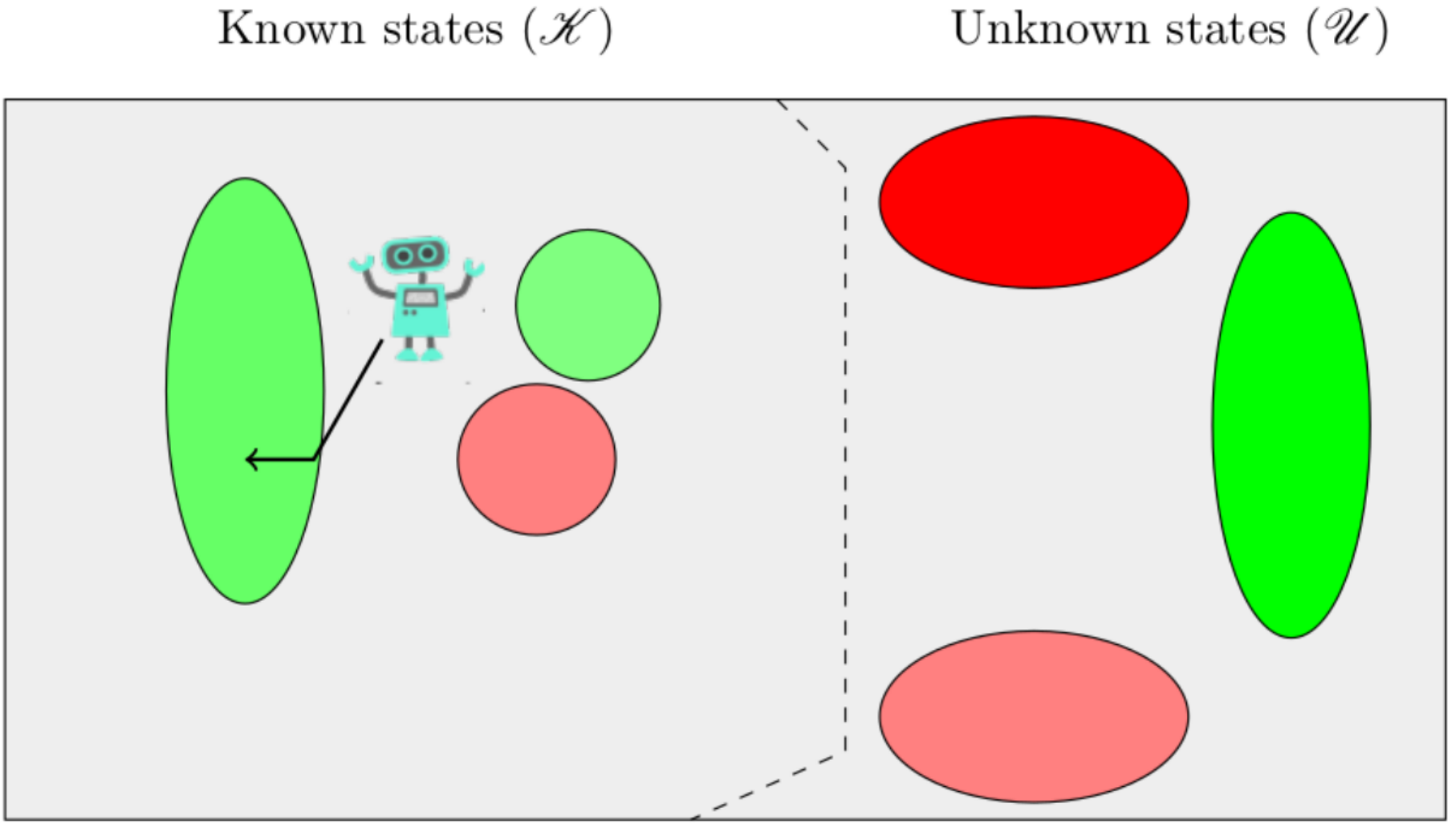}} \quad  \subfigure[Explore]{\includegraphics[height=0.13\textheight]{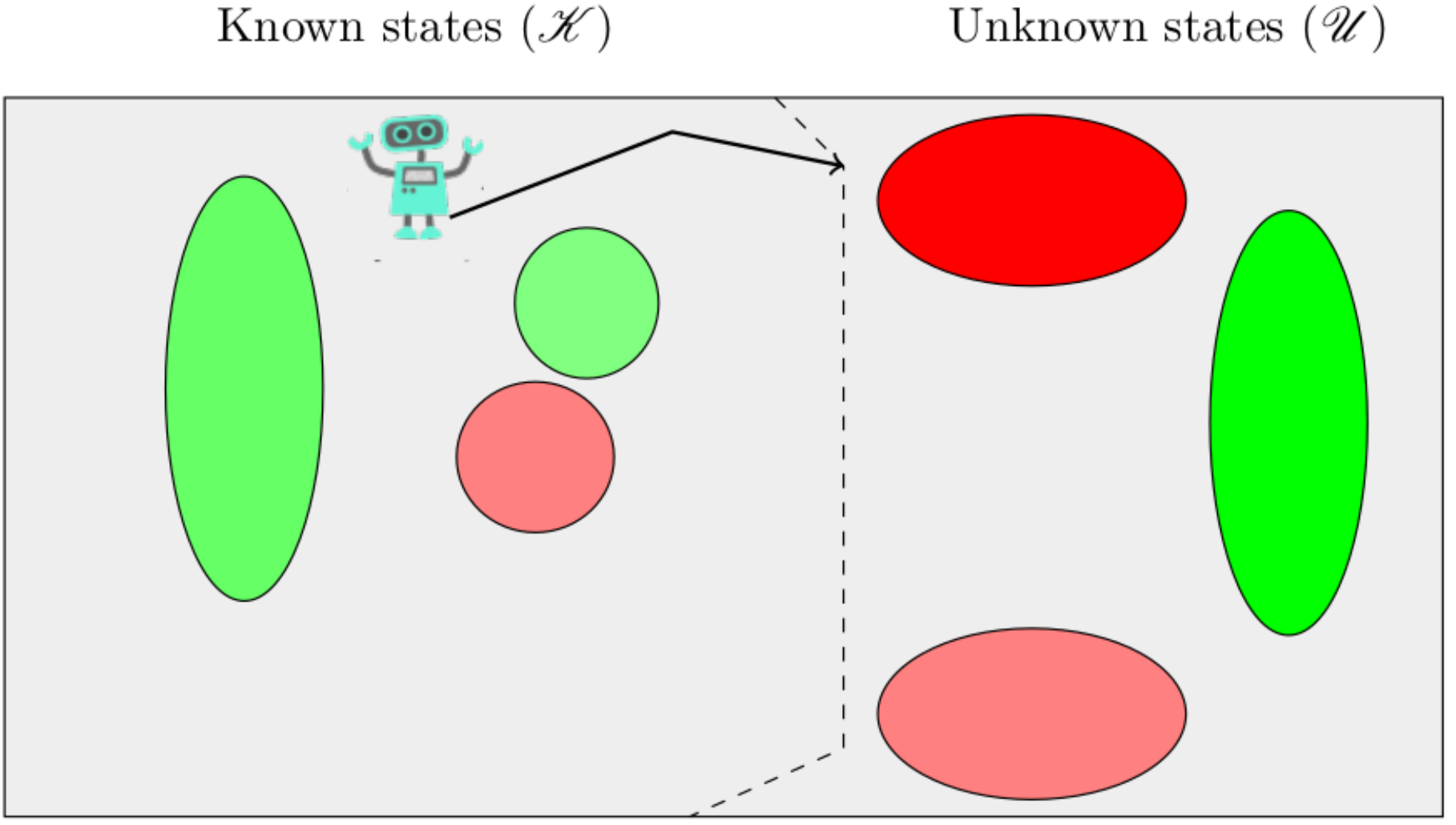}}\\
\hspace*{0.13\textwidth} \subfigure[Escape]{\includegraphics[height=0.13\textheight]{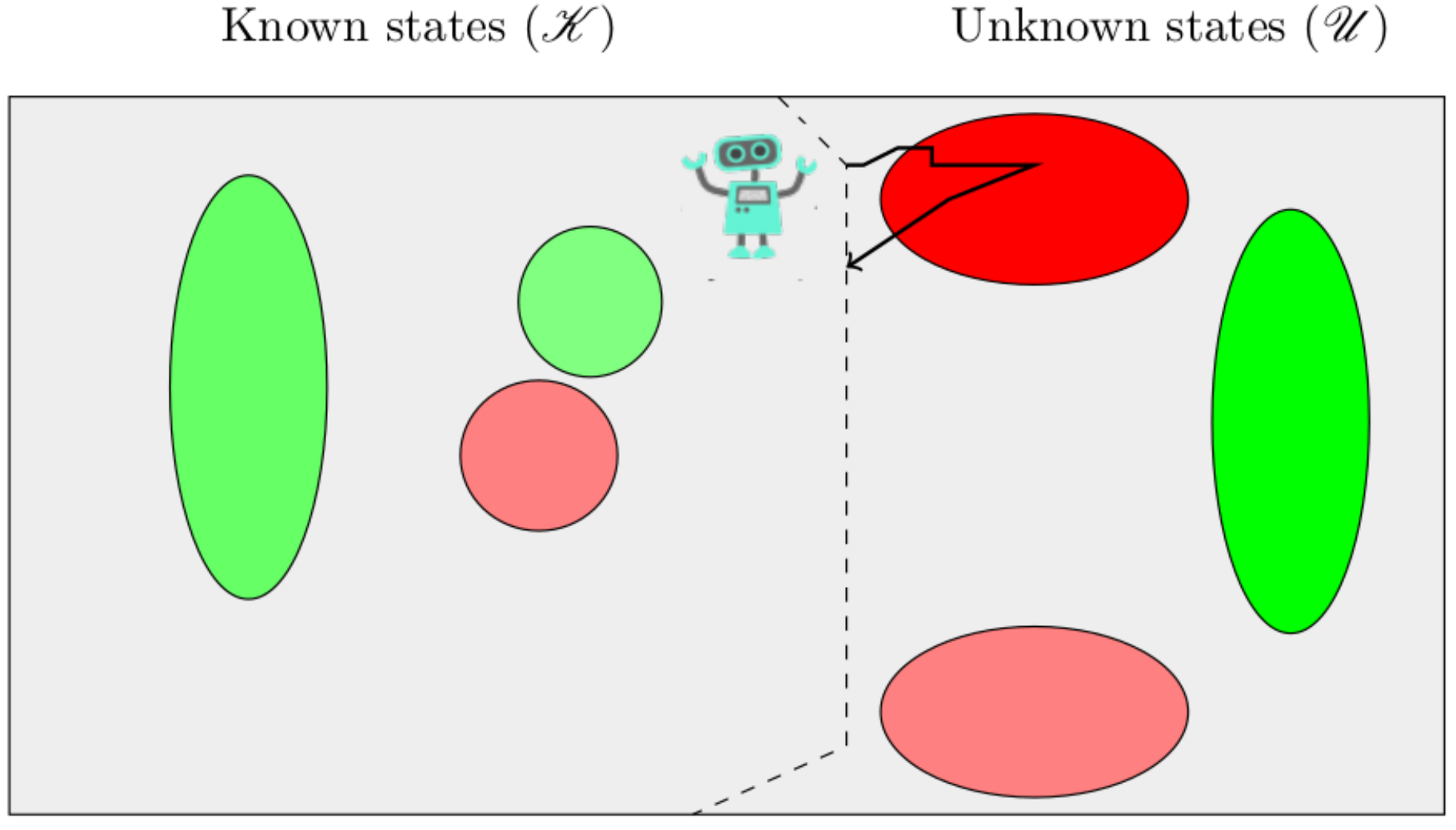}} \quad  \subfigure[Exploit (End)]{\includegraphics[height=0.13\textheight]{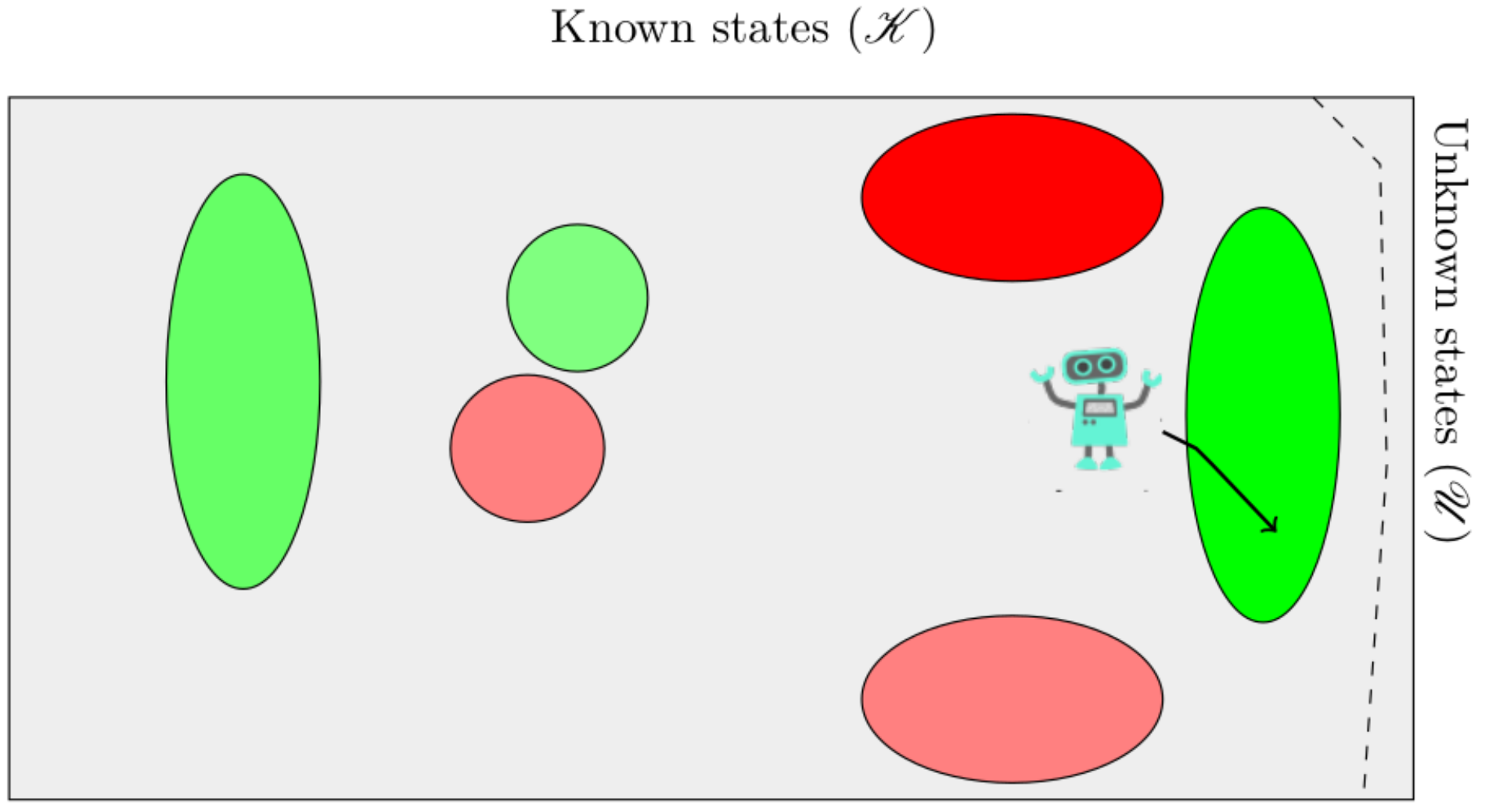}}
\caption{Diagram illustrating $E^4$. $E^4$ repeatedly takes limited-step trajectories in either known or unknown states. \textbf{(a)} The ``exploitation policy'' has the aim of solving the CMDP within the known states. If from a known state it has near-optimal constrained limited-step performance, then this policy is applied. \textbf{(b)} From another state, the exploitation policy may not be near-optimal; in this case an ``exploration policy'' aims to reach the unknown states, in an attempt to make them known. \textbf{(c)} Once in the unknown states, a balanced wandering behaviour ensures that actions are equally often taken for each unknown state, thereby yielding reliable statistics for different actions. Once there is an observed future risk of constraint-violation, an ``escape policy'' goes back to the known states as quickly as possible. \textbf{(d)} At the final stages of the $E^4$ algorithm, nearly all states have been frequently visited with balanced action-visitations, making them known. This then allows finding a near-optimal exploitation policy for the full CMDP.}\label{fig: diagram}
\end{figure}

What is currently missing from $E^3$ is a suitable way to deal with constraint-satisfaction across the known and unknown states, as well as a suitable method for providing a safe return, or ``escape'', from the unknown states back to the known states. Considering an additional ``escape'' policy in addition to an exploration and exploitation policy, this paper formulates an algorithm, called Explicit Explore, Exploit, or Escape ($E^4$; see Figure~\ref{fig: diagram}). $E^4$ extends the $E^3$ algorithm to satisfy safety constraints throughout the lifetime of the RL agent through the following algorithmic contributions.
\begin{itemize}
\item Using the CMDP rather than an MDP framework, $E^4$ additionally models another stream of reinforcement signals called constraint-costs, which are constrained to a long-term budget and which represent the cost of state-action pairs.
\item A correction is formulated for the constraint-cost budget in offline optimisation to account for the potential model errors in known states.
\item Using a specialised ``escape policy'', optimised to return to the known states as quickly as possible, $E^4$ halts the balanced wandering behaviour of $E^3$ as soon as there is a risk of exceeding the budgetary constraint in unknown states.
\item Analytical formulas are derived to determine an allowed pseudo-budget in known and unknown states to ensure the budget over the full CMDP is not exceeded.
\item To ensure the escape policy returns without exceeding the pseudo-budget in unknown states, even under worst-case assumptions, we propose four possible approaches, making modifications to existing robust and constrained optimisation algorithms:
\begin{itemize}
\item using the robust CMDP policy gradient \citep{Russel2020} as is in unknown states and optionally in known states;
\item using the robust linear programming technique by \cite{Zheng2020}, which is applicable to episodic and undiscounted CMDPs with known transition dynamics, in the known states, by accounting for the discount factor and by considering a limited-step trajectory within the full lifetime of the $E^4$ agent;
\item incorporating an uncertainty over transition dynamics into the constraints of constrained DP  \citep{Altman1998}; and
\item incorporating constraints into robust DP \citep{Nilim2005,Iyengar2005} by reformulating the CMDP as a Lagrangian MDP \citep{Taleghan2018}.
\end{itemize}  
\end{itemize}
Applying these algorithmic principles, we demonstrate that $E^4$ finds a near-optimal policy within the set of policies that satisifies the budgetary constraints.
Doing so yields similar sample complexity as $E^3$ but adds time complexity due to robust optimisation. Compared to the works mentioned above, the key distinctive features of $E^4$ are 1) safety-constraint satisfaction throughout the lifetime; 2) the explicit explore, exploit, or escape structure;  3) modelling the transition dynamics, reward function, constraint-cost function, and their uncertainty for robust offline optimisation; and 4) the aim to find a near-optimal constrained policy, rather than to improve on a known policy.

\section{Preliminaries and definitions}
\label{sec: definitions}
The framework for $E^4$ is based on the CMDP task-modelling framework. In CMDPs, at each time step, an agent receives a state, performs an action, receives a reward, and a constraint-cost; the goal of the agent is to maximise the long-term cumulative reward whilst not exceeding a pre-defined budget of constraint-cost. Formally, a CMDP is a tuple $\langle \mathcal{S}, \mathcal{A}, P^{*}, r, c, d, \gamma \rangle$, where $\mathcal{S}
$ is the finite state space of size $S$; $\mathcal{A}$ is the finite action space of size $A$; $P^{*} : \mathcal{S} \times \mathcal{A}  \to \Delta^S$ is the true transition dynamics of the environment with $\Delta^S = \{ p \in \mathbb{R}^S : p^T \mathbf{1} = 1  \}$ being the probability simplex in $\mathcal{S}$; $r(s,a)$ is the average reward obtained for performing action $a \in \mathcal{A}$ in state $s \in 
\mathcal{S}$; $c(s,a)$ is the average constraint-cost for performing action $a \in \mathcal{A}$ state $s \in \mathcal{S}$; $d$ is the budget on the expected asymptotic constraint-cost (see below); and $\gamma 
\in [0,1)$ is the discount factor used for computing the long-term value and constraint-cost function based on the expected cumulative reward and constraint-cost, 
respectively. When at time $t$ the RL agent is presented with a state $s_t$, its objective within the CMDP is to find a policy $\pi : \mathcal{S} \to \mathcal{A}$ that 
maximises the expected asymptotic value,
\begin{equation}
V_{\pi}(s_t) = \mathbb{E}_{\pi,P^{*}} \left[ \sum_{k=0}^{\infty} \gamma^{k} r(s_{t+k},a_{t+k}) \right] \,,
\end{equation}
while the expected asymptotic constraint-cost, defined as
\begin{equation}
C_{\pi}(s_t) = \mathbb{E}_{\pi,P^{*}} \left[ \sum_{k=0}^{\infty} \gamma^{k} c(s_{t+k},a_{t+k}) \right] \,,
\end{equation}
must satisfy $C_{\pi}(s_t) \leq d$. Moreover, assumptions on the CMDP are three-fold. First, the reward distribution for a state-action pair $(s,a) \in \mathcal{S} \times \mathcal{A}$ has mean $r(s,a)  \in [0,r_{\text{max}}]$ and 
variance $\text{Var}(s) \leq \text{Var}_{\text{max}}^r$. Second, the constraint-cost distribution for a state-action pair $(s,a) \in \mathcal{S} \times \mathcal{A}$ has a mean $c(s,a)  \in [0,c_{\text{max}}]$ and variance 
$\text{Var}^c(s) \leq \text{Var}_{\text{max}}^c$. Third, a general upper-bound for the expected T-step value is denoted by $G_{\text{max}}^r(T) = \sum_{t=0}^{\infty} \gamma^{t} r_{\text{max}} = \frac{r_{\text{max}}}{1-\gamma}$ while analogously the general upper-bound for the expected T-step constraint-cost is denoted by $G_{\text{max}}^c(T) = \sum_{t=0}^{\infty} \gamma^{t} c_{\text{max}} = \frac{c_{\text{max}}}{1-\gamma}$. In practice, the above-mentioned maximal variance, reward, value, etc. do not need to be known exactly; any upper bound suffices, although tighter bounds improve the results. Knowledge of $r_{\text{max}}$ and $c_{\text{max}}$ are a basic requirement since the other maxima can be upper-bounded from them. To give a few examples of when $r_{\text{max}}$ and $c_{\text{max}}$ can be known exactly or upper-bounded, one can consider applications where rewards or constraint-costs represent a limited resource, such as energy, food, or money, or a physical force, such as friction or torque. A further property of the CMDP being studied is that it has a limited ``diameter''. Completely analogous to the diameter \citep{Jaksch2010} for MDPs, we define the diameter of a CMDP $M$ as the maximal expected number of actions from one arbitrary state $s \in \mathcal{S}$ to any other $s' \in \mathcal{S}$ under the best possible stationary deterministic policy $\pi: \mathcal{S} \to \mathcal{A}$ for the choice of $(s,s')$, or
\begin{equation}
D(M) := \max_{s \neq s'} \min_{\pi} \mathbb{E}\left[ W(s' \vert s, M, \pi) \right] \,,
\end{equation}
where $W(s' \vert s, M, \pi)$ is the number of actions from $s$ to $s'$ given the model $M$ and policy $\pi$.

While the above functions span over an infinite horizon, it is often of interest to compute their $T$-
step approximation to make decisions based on limited data. The corresponding $T$-step approximations instead sum over $T$ steps and are denoted by  $V_{\pi}(s_t,T)$ 
and $C_{\pi}(s_t,T)$, respectively. From the class of policies $\Pi$, the class of constrained policies for any state $s \in \mathcal{S}$, a budget $d$, a number of time steps $T$, is denoted as $\Pi_c(s,d,T) = 
\{\pi \in \Pi: C_{\pi}(s,T) \leq d   \}$. Denoting $\Pi_c(s,d)$ for $T\to\infty$, $\Pi_c(s,d) \neq \emptyset$ is non-empty by assumption, and therefore any other choice of $T$ yields $\Pi_c(s,d,T) \neq \emptyset$. An optimal constrained policy for a T-step trajectory in $M$ is then defined as $\pi^{*} = \argmax_{\pi \in \Pi(s,d,T)} V_{\pi}
(s,T)$ where for the asymptotic case of course $T \to \infty$. In addition to the expected asymptotic constraint-cost, which computes the expectation over different possible trajectories, the analysis also uses a path-based constraint-cost and value, which computes the average cost for a particular state-action trajectory and is denoted for a $T$-step path $p$ as $C(p) = \sum_{t=0}^{T-1} \gamma^t c(s_t,a_t)$  and $V(p) = \sum_{t=0}^{T-1} \gamma^t r(s_t,a_t)$, respectively. 

With $M$ being the CMDP to be solved, this paper takes a model-based approach in which an approximate model $\hat{M} = \langle \mathcal{S}, \mathcal{A}, \hat{P}, 
\hat{r}, \hat{c}, d', \gamma \rangle$ is continually improved by sampling from $M$, with hat-notations emphasising the model and its components are estimated by the sample mean, and where a potentially lower budget $d'$, $c_{\text{max}} \leq d' \leq d$, is used when the model is induced over a subset of the full CMDP (see Section \ref{sec: inducedCMDP}), where the relation with $c_{\text{max}}$ is an assumption of the algorithmic approach to ensure at least one action can be performed before potentially exceeding $d'$. For states that have not yet been visited frequently, we formulate an
uncertainty set (or ambiguity set), $\mathcal{P}$, as the set of transition dynamics models consistent with the samples from (or prior knowledge on) the true but unknown transition dynamics, $P^{*}$. For example, one can define a confidence interval around $\hat{P}_{s,a}$, the estimated (also called ``nominal'') transition dynamics model, with a budget $\psi_{s,a}$ for state-action pair $(s,a)$:
\begin{equation}
\mathcal{P}_{s,a} = \{ P \in  \Delta^{S} : \vert\vert P - \hat{P}_{s,a} \vert\vert_1 \leq \psi_{s,a}    \} \,.
\end{equation}
Within the states that have not been frequently visited yet, the uncertainty set is then used to optimise the constraints robustly, that is, with $\max_{P \in \mathcal{P}} \hat{C}_{\pi}(s_t) \leq d$, where $\hat{C}_{\pi}$ indicates the estimate of the expected asymptotic constraint-cost, with more precise definitions and notations to follow. 

\section{Main theorem}
\label{sec: theory}
With the preliminary definitions in mind, this section states and proves the main theorem, which postulates the existence of an algorithm, namely $E^4$, which explores a CMDP safely within the constraint-cost budget and which finds a near-optimal constrained policy within polynomial time.
\begin{theorem}
\label{theorem1}
There exists an algorithm ($E^4$) that outputs with probability $1 - \delta$ a near-optimal constrained policy $\pi$ for $s$ with $V_{\pi}(s) \geq V_{\pi^{*}}(s) - \epsilon$ and $C_{\pi}(s) \leq d$ with sample complexity and time complexity that is polynomial in $1/\epsilon$, $1/\delta$, $S$, the horizon time $1/(1-\gamma)$,  $r_{\text{max}}$, and $c_{\text{max}}$. Moreover, the non-stationary policy $\pi^n=\{\pi_{i}\}_{i=1}^{\infty}$ induced by the exploration process of $E^4$ yields expected asymptotic constraint-cost $C_{\pi^n}(s_t) \leq d$ with probability at least $1 - (UA\delta_{\psi} + \delta)$.
\end{theorem}

The $E^4$ algorithm is based on the model-based CMDP framework presented in the previous section. The Constrained Simulation Lemma (Section~\ref{sec: CS-lemma}) shows that a sufficiently accurate CMDP model of the reward function, constraint-cost function, and transition dynamics allow, with high probability, an $\epsilon$-correct value function and constraint-cost function approximation. Moreover, it is sufficient to visit each state-action pair a number of times that is polynomial in $S$, $T$, $1/(1-\gamma)$, $r_{\text{max}}$,$1/\epsilon$, and $1/\delta$ to obtain such an accurate CMDP model. This leads to a natural definition of ``known states'', which have been visited often with high frequency for each action and thereby have accurate models, and ``unknown states'', which have not been visited often enough. Each of the two cases can then be treated as a different Induced CMDP (see Section~\ref{sec: inducedCMDP}). Therefore when starting a limited $T$-step trajectory from a state $s \in \mathcal{S}$, its strategy for known states differs from that in the unknown states. 

Starting from a known state, the agent may already have a near-optimal policy available. If so, then this policy is best exploited for the following $T$-step trajectory. If not, then the model must be further improved by exploring unknown states.  This is formalised by the $l$-safe Explore-or-Exploit Lemma (Section~\ref{sec: CEE-lemma}). This lemma shows that for policies satisfying a given a constraint-budget within the known states, there either exists a ``exploitation policy'' which is a near-optimal constrained policy or there exists an ``exploration policy'' which with high probability finds an unknown state. The $l$-safe Explore-or-Exploit Lemma then shows how to define a correction to the original budget to account for the approximation error of the model, resulting in a high-probability guarantee for satisfaction of the original constraint across the known states. 

When exploring unknown states, the agent must attempt to make these states known by performing many state-action visitations. A key challenge is that this risks exceeding the constraint-cost budget because the agent has little knowledge of the unkown states. With worst-case assumptions on the transition dynamics and constraint-costs, the Safe Escape Lemma (Section~\ref{sec: SE-lemma}) provides a high-probability guarantee for an optimised ``escape policy'' to find a path from the unknown states to the known states whilst satisfying the constraint-cost budget defined in the unknown states. Since the escape policy will not select each action equally frequently from a given state, the escape policy provides no guarantee on making states known. To safely gain as much knowledge on the unknown states as possible, $E^4$ performs a balanced wandering behaviour, which selects the least-taken action for a given state, as long as timely escape is ensured by the escape policy, as formalised by the Safe Balanced Wandering Lemma (Section \ref{sec: safe balanced wandering}).

To integrate the results for the known and unknown states, the Escape Budget Lemma (Section~\ref{sec: EB-lemma}) shows how to set the budget for the unknown states, $d'$, to satisfy the budget on the full CMDP ($d$) given a budget within the known states $d''$; this also helps to define the conditions for constraint-satisfiability. The remainder of the section (Section~\ref{sec: together}) then puts all the lemmas together and demonstrates the validity of Theorem \ref{theorem1}.

\subsection{Constrained Simulation Lemma}
\label{sec: CS-lemma}
Since the model is estimated based on the sample mean, repeatedly visiting the same state allows estimating the transition dynamics, the reward function, and the constraint-cost function. Therefore, the true CMDP $M$ can be approximated by simulation model $\hat{M}$ with a sufficient number of samples. If the simulation model is sufficient, the expected asymptotic value and constraint-cost of the CMDP can be approximated with high accuracy, a finding formalised by the Constrained Simulation Lemma (Section~\ref{lem: CS}). The set of ``known states'' is then defined as those states which have sufficient visitations and are therefore approximated correctly.
   
The Constrained Simulation Lemma relies on the relation between the expected asymptotic and T-step value and constraint-cost. Specifically, the following Lemma shows that for sufficiently large $T$, the expected asymptotic value and constraint-costs are $\epsilon$-close to their $T$-step counterparts.
\begin{lemma}
\label{lem: CTE}
\textbf{Constrained T-step Estimation Lemma:} Let $M$ be the CMDP, $\pi$ be a policy in $M$, and $T \geq \frac{1}{1-\gamma} \ln\left(\frac{\max(r_{\text{max}},c_{\text{max}})}{\epsilon (1-\gamma)}  \right)$. Then for any state $s \in \mathcal{S}$ we have \\
\textbf{(a)} $V_{\pi}(s,T) \leq V_{\pi}(s) \leq V_{\pi}(s,T) + \epsilon$; and \\ 
\textbf{(b)} $C_{\pi}(s,T) \leq C_{\pi}(s) \leq C_{\pi}(s,T) + \epsilon$.\\
For these reasons, we call $\frac{1}{1-\gamma} \ln\left(\frac{\max(r_{\text{max}},c_{\text{max}})}{\epsilon (1-\gamma)}\right)$ the \textbf{$\epsilon$-horizon time of the CMDP $M$}.
\end{lemma}
\textbf{Proof of (a):}  Define MDP $M^- = \langle \mathcal{S},\mathcal{A}, r, \gamma, P \rangle$ with the same states, action space, reward function, discount factor, and transition dynamics as $M$. Let $\pi$ be any policy in $M^-$. The result follows directly from the original T-step estimation Lemma (see Lemma 2 in \cite{Kearns2002}) as this holds for any policy in any MDP. In short, the first inequality $
V_{\pi}(s,T) \leq V_{\pi}(s)$ follows from rewards being non-negative and the expected asymptotic value additionally considering time steps $t= T+1, \dots, \infty$;  the second inequality follows after requiring $\epsilon \leq \gamma^T \frac{r_{\text{max}}}{1-\gamma}$ (the maximal remaining reward not accounted for after time step $T$) and then solving for $T$.  \\
\textbf{Proof of (b):}
The proof is completely analogous as the transition dynamics and policy are the same while the constraint-cost function is analogously bounded in $[0,c_{\text{max}}]$. \qed

The basic reasoning behind the Constrained Simulation Lemma is that, if an estimator $\hat{M}$ of the true CMDP $M$ is sufficiently accurate, then $\hat{M}$ can be used as a simulation model to simulate realistic trajectories of $M$; with offline optimisation algorithms, this can then yield nearly correct estimates of the expected asymptotic value and the expected asymptotic constraint-cost. To develop this reasoning, a suitable definition for the accuracy of such a simulation model is formalised below.
\begin{definition} 
CMDP $\hat{M}$ is an \textbf{$\alpha$-approximation} of CMDP $M$ if and only if:\\
1) for all state-action-pairs $(s,a) \in \mathcal{S} \times \mathcal{A}$ and : $r(s,a) - \alpha \leq \hat{r}(s,a) \leq r(s,a) + \alpha$; \\
2) for all transitions $(s,a,s') \in \mathcal{S} \times \mathcal{A} \times \mathcal{S}$: $P^{*}_{s,a}(s') - \epsilon \leq \hat{P}_{s,a}(s') \leq P^{*}_{s,a}(s') + \epsilon$.\\
3) for all state-action-pairs $(s,a) \in \mathcal{S} \times \mathcal{A}$: $c(s,a) - \alpha \leq \hat{c}(s,a) \leq c(s,a) + \alpha$. \\
\end{definition}

With this definition in mind, the Constrained Simulation Lemma states that as long as $T$ is chosen according to the $\epsilon$-horizon time of $M$ and the simulation $\hat{M}$ is $\alpha$-correct, the simulated value and constraint-costs are $\epsilon$-correct with respect to the true expected asymptotic value and constraint-cost.
\begin{lemma}
\label{lem: CS}
\textbf{Constrained Simulation Lemma.} Let $T \geq \frac{1}{1 - \gamma} \ln(\frac{\max(r_{\text{max}},c_{\text{max}})}{\epsilon(1 - \gamma)})$, let CMDP $\hat{M}$ be an $\alpha$-approximation of CMDP $M$ with $\alpha=O\left(\left( \epsilon / \left(STG \right)\right)^2\right)$ where $G=\max(G_{\text{max}}^r(T),G_{\text{max}}^c(T))$, and let $\pi$ be a policy in $M$. Then for any state $s$, \\
\textbf{(a)} $V_{\pi}(s) - \epsilon \leq \hat{V}_{\pi}(s) \leq V_{\pi}(s) + \epsilon$; and \\
\textbf{(b)} $C_{\pi}(s) - \epsilon \leq \hat{C}_{\pi}(s) \leq C_{\pi}(s)+ \epsilon$.
\end{lemma}
The proof extends the original Simulation Lemma (see Lemma 4 in \cite{Kearns2002}) to include the constraint-costs. As it is extensive and otherwise analogous, it is provided in the appendix.

Having established the relation between the expected asymptotic constraint-cost and value on the one hand and the $\alpha$-approximation on the other hand, the following lemma provides the number of samples for an accurate estimate of the expected asymptotic constraint-cost and value with error of at most $\epsilon$.
\begin{lemma}
\label{lem: KS}
\textbf{Known State Lemma} Given state $s$ has been visited 
\begin{equation}
\label{eq: m}
m = O((STG/\epsilon)^4 \text{Var}_{\text{max}} \ln(1/\delta))
\end{equation}
times, where $G=\max(G_{\text{max}}^r(T),G_{\text{max}}^c(T))$ and $\text{Var}_{\text{max}}=\max(\text{Var}_{\text{max}}^r,\text{Var}_{\text{max}}^c)$, and from $s$ each action has been executed at least $\lfloor m / A \rfloor$ times, then with probability of at least $1 - \delta$ we have\\
\textbf{a)} $\vert P^{*}_{s,a}(s') - \hat{P}_{s,a}(s')\vert = O(\epsilon/(STG))^2)$ for any $(s,a,s') \in \mathcal{S} \times \mathcal{A} \times \mathcal{S}$; \\
\textbf{b)} $\vert r(s,a) - \hat{r}(s,a)\vert = O((\epsilon/(STG))^2)$ for any $(s,a) \in \mathcal{S} \times \mathcal{A}$; and \\
\textbf{c)} $\vert c(s,a) - \hat{c}(s,a)\vert  = O((\epsilon/(STG))^2)$ for any $(s,a) \in \mathcal{S} \times \mathcal{A}$.\\
\end{lemma}
\textbf{Proof:}\\
The proof is based on Hoeffding's inequality and Chernoff bounds (see Appendix A for details).\qed

Following the above lemma, the notion of ``known states'' can now be defined directly from Eq.~\ref{eq: m}.  Intuitively, contrasting known versus unknown states allows  to distinguish regions for which expected asymptotic value and constraint-cost is estimated correctly, and regions for which they are not. This contrast between known and unknown states will be critical to provide safety guarantees for the RL algorithm later on.
\begin{definition}
\label{def: KS}
\textbf{Known states $\mathscr{K}$ and unknown states $\mathscr{U}$:} a state $s \in \mathcal{S}$ is called \textbf{known} if $\forall a \in \mathcal{A}: n(s,a) \geq m_{\text{known}}$ where $m_{\text{known}} \in \mathbb{N}$ is $O((STG/\epsilon)^4 \text{Var}_{\text{max}} \ln(1/\delta))$ following Eq.~\ref{eq: m}. Alternatively, $s \in \mathcal{S}$ is called \textbf{unknown} if $\exists a \in \mathcal{A}: n(s,a) < m_{\text{known}}$. Known states are collectively referred to as $\mathscr{K} = \{s \in \mathcal{S} \vert \forall a \in \mathcal{A}: n(s,a) \geq m_{\text{known}}\}$ while unknown states are referred to as $\mathscr{U} = \{s \in \mathcal{S} \vert \exists a \in \mathcal{A}: n(s,a) < m_{\text{known}}\}$.
\end{definition}

\subsection{Induced CMDPs}
\label{sec: inducedCMDP}
Since only known states are modelled to a sufficient accuracy, it is useful to consider CMDP simulations over the known states only. The following defines the notion of an ``induced CMDP'', which limits the original CMDP to a subset of the original state space (e.g., the known states). The induced CMDPs allow two useful results. First, since the induced CMDP is also a CMDP like in the Constrained Simulation Lemma, an analogous result holds for induced CMDPs over the set of known states (see Lemma~\ref{lem: IS}); that is, an estimated induced CMDP is $\epsilon$-correct with probability at least $1 - \delta$ when compared to the true induced CMDP. Second, the expected $T$-step value and constraint-cost within induced CMDPs provide lower bounds for the expected $T$-step value and constraint-cost, respectively, in the original CMDP (see Lemma~\ref{lem: IU}). 

\begin{definition}
\label{def: ICMDP}
Given a CMDP $M$, define for a subset of states $\mathscr{S} \subset \mathcal{S}$ an \textbf{induced CMDP} $M_{\mathscr{S}}$, which is equal in all but the following respects:
\begin{itemize}
\item its state space is $\mathscr{S} \cup s_0$ where $s_0$ is a terminal state of the CMDP.
\item transitions to states in $s \not\in \mathscr{S}$ are redirected to a terminal state $s_0$, which terminates the episode and yields terminal reward $r(s_0)=0$ and terminal constraint-cost $c(s_0)=0$.
\item for each $(s,a) \in \mathcal{S}$, rewards and constraint-costs always yield their mean $r(s,a)$ and $c(s,a)$ deterministically, with zero variance.
\end{itemize}
When a CMDP is induced over the known states $\mathscr{K}$ (see Definition 2), the resulting CMDP is denoted as $M_{\mathscr{K}}$ and is called the \textbf{known-state CMDP} of $M$.
\end{definition}
In the following, whenever a value function, constraint-cost, or other quantity is computed in an induced CMDP $M_{\mathscr{S}}$, this will be denoted by the conditioning operator $\vert M_{\mathscr{S}}$; for example, $V_{\pi}(s \vert M_{\mathscr{K}})$ denotes the expected asymptotic value of $\pi$ in the known state CMDP. For brevity and consistency with the earlier sections, when the conditioning is on the full CMDP of interest, $M$, the conditioning is omitted; for example, $V_{\pi}(s)$ denotes the expected asymptotic value of $\pi$ in the full CMDP of interest.

Now the so-called Induced Simulation Lemma applies the Constrained Simulation Lemma to known-state CMDPs.
\begin{lemma}
\label{lem: IS}
\textbf{Induced Simulation Lemma} Let $M$ be the CMDP of interest and let $M_{\mathscr{K}}$ be its known state CMDP, and let $T \geq \frac{1}{1-\gamma} \ln\left(\frac{\max(r_{\text{max}},c_{\text{max}})}{\epsilon (1-\gamma)}  \right)$. Then, for any policy $\pi$ and any state $s \in \mathcal{S}$, with probability $1 - \delta$, we have
\begin{align}
V_{\pi}(s \vert M_{\mathscr{K}}) - \epsilon \leq \hat{V}_{\pi}(s \vert \hat{M}_{\mathscr{K}})   \leq V_{\pi}(s \vert M_{\mathscr{K}}) + \epsilon \nonumber
\end{align}
and 
\begin{align}
C_{\pi}(s \vert M_{\mathscr{K}}) - \epsilon \leq \hat{C}_{\pi}(s \vert \hat{M}_{\mathscr{K}})   \leq C_{\pi}(s \vert M_{\mathscr{K}}) + \epsilon \,. \nonumber
\end{align}
\end{lemma}
\textbf{Proof:}\\
The proof follows directly from the Constrained Simulation Lemma (see Lemma~\ref{lem: CS}) and the Known State Lemma (see Lemma~\ref{lem: KS}). \qed

As one is interested in what happens in the ``real world'', the below lemma relates the induced CMDP to the original CMDP $M$.
\begin{lemma}
\label{lem: IU}
\textbf{Induced underestimation lemma:} Let $\mathscr{S} \subset \mathcal{S}$. For any $s \in \mathscr{S}$, and any policy $\pi$ in $M$, we have
\begin{align}
V_{\pi}(s, T \vert M_{\mathscr{S}}) \leq V_{\pi}(s, T) \nonumber
\end{align}
and 
\begin{align}
C_{\pi}(s, T \vert M_{\mathscr{S}}) \leq C_{\pi}(s, T) \,. \nonumber
\end{align}
\end{lemma}
\textbf{Proof:} Let $\pi$  be a policy in $M$ and let $s \in \mathscr{S}$. Then we have, analogous to the original lemma, the following results because $\pi$ stops at time $T_{\text{stop}} \leq T$ and rewards and constraint-costs are positive. To illustrate, for the value function, we have:
\begin{align}
V_{\pi}(s, T \vert M_{\mathscr{S}})= \sum_p \mathbb{P}_{\pi,P^{*}}[p ] \sum_{t=0}^{T_{\text{stop}}-1} \gamma^{t} r_t \leq V_{\pi}(s, T)  \nonumber 
\end{align}
due to rewards being positive and the paths being terminated earlier at a time $T_{\text{stop}} \leq T$ at terminal state $s_{T_{\text{stop}}} \in \mathcal{S} \setminus \mathscr{S}$. The result for the constraint-cost is completely analogous.
\qed

\subsection{l-safe Explore-or-Exploit lemma}
\label{sec: CEE-lemma}
Following the definition of induced CMDPs, this section provides results for exploration and exploitation within the known states. For any induced CMDP which has satisfiable constraint-cost budget, the Constrained Explore-or-Exploit lemma (Lemma~\ref{lem: CEE}) proves the existence of either an exploitation policy, which solves the CMDP within the known states near-optimally,  and an exploration policy, which gives a high-probability guarantee of visiting an ``unknown state''. This allows the $l$-safe Explore-or-Exploit Lemma (Lemma~\ref{lem: lEE}), which provides safety, i.e. guarantees on constraint-satisfaction, within the known states by defining a lower budget $l$ that accounts for worst-case modelling errors (hence the term $l$-safe). The budget $l$ subtracts the model errors from Lemma~\ref{lem: CTE} and Lemma~\ref{lem: IS} to the Constrained Explore-or-Exploit Lemma. 

Based on defining an induced CMDP over a subset of states $\mathscr{S}$ (e.g. the known states, in which case $\mathscr{S}=\mathscr{K}$), the Constrained Explore-or-Exploit Lemma proves either the existence of an optimal constrained policy on the full CMDP or a constrained policy that finds a state not in $\mathscr{S}$ in $T$-steps with high probability.
\begin{lemma}
\label{lem: CEE}
\textbf{Constrained Explore-or-Exploit Lemma} 
Let $M$ be any CMDP, let $\mathscr{S}$ be any subset of states $\mathscr{S} \subset \mathcal{S}$, and let $M_{\mathscr{S}}$ be the induced CMDP over $\mathscr{S}$ with a given budget $d$. For any $s \in \mathscr{S}$, for any $T$, and any $\epsilon \geq 0$, we have \textbf{either a)} there exists a policy $\pi \in M_{\mathscr{S}}$ for which $V_{\pi}(s, T \vert M_{\mathscr{S}}) \geq V_{\pi^{*}}(s, T) - \epsilon$, where $\pi^{*} = \argmax_{\pi \in \Pi_c(s,d,T)} V_{\pi}(s,T)$ is the optimal constrained $T$-step policy, and which satisfies $C_{\pi}(s, T \vert M_{\mathscr{S}}) \leq d$; \textbf{or b)} there exists a policy $\pi$ in  $M_{\mathscr{S}}$ which reaches the terminal state $s_0$ in $\mathcal{S} \setminus \mathscr{S}$ in at most $T$ steps with probability $p> \epsilon /G_{\text{max}}^r(T)$, and $C_{\pi}(s, T \vert M_{\mathscr{S}}) \leq d$.
\end{lemma}
\textbf{Proof: } \\
The proof and lemma is analogous to the original Explore-or-Exploit Lemma, except that the optimal constrained policy rather than the optimal policy is considered (see Appendix B).\\
\qed

Now by integrating the Constrained Simulation Lemma and the Constrained Explore-or-Exploit Lemma, a characterisation of the safety can be provided in terms of the probability of constraint-satisfaction during exploitation or exploitation in the known-state CMDP. This safety guarantee,  based on the \textit{estimated} known-state CMDP, is provided by the $l$-safe Explore-or-Exploit Lemma below.
\begin{lemma}
\label{lem: lEE}
\textbf{$l$-safe Explore-or-Exploit Lemma:} Let $\epsilon \in [0,d/2)$, $l \leq d - 2\epsilon$, $\delta \in [0,1]$, $\mathscr{K}$, and $T  \geq \ln\left(\frac{\max(r_{\text{max}},c_{\text{max}})}{\epsilon(1-\gamma)}\right) \frac{1}{1-\gamma}$ (i.e. at least the $\epsilon$-horizon time). Furthermore, let $M_{\mathscr{K}}$ be the known-state CMDP and $\hat{M}_{\mathscr{K}}$ its estimate. Then with probability $1- \delta$, a policy $\pi$ that satisfies $\hat{C}_{\pi}(s, T\vert \hat{M}_{\mathscr{K}}) \leq l$ for $s \in \mathscr{K}$ will satisfy $C_{\pi}(s \vert M_{\mathscr{K}}) \leq d$.
\end{lemma}
\textbf{Proof:} \\
Due to the Induced Simulation Lemma and the Constrained T-step Estimation Lemma, we have for any state $s \in \mathcal{S}$ with probability $1 - \delta$:
\begin{align}
C_{\pi}(s \vert M_{\mathscr{K}})  \leq \hat{C}_{\pi}(s\vert \hat{M}_{\mathscr{K}}) + \epsilon \leq &\hat{C}_{\pi}(s, T\vert \hat{M}_{\mathscr{K}}) + 2\epsilon  \leq l + 2\epsilon \leq d \,. \nonumber
\end{align}
 \qed

\subsection{Safe escape lemma}
\label{sec: SE-lemma}
Contrasting to the known states, any model based on the unknown states is inaccurate, making it difficult to guarantee safe return to the known states before exceeding the constraint-cost budget. To overcome the challenge, we construct a ``Worst-case  Escape CMDP'', a CMDP induced over the unknown states that rewards escaping back to the known states before exceeding a constraint-cost budget and makes worst-case assumptions on the constraint-cost. Additionally incorporating worst-case transition dynamics through the use of an uncertainty set, the resulting robust CMDP provides a probabilistic guarantee for safety within the unknown states. The proof relies on robust constraint-satisfiability, a condition that is further discussed in Section~\ref{sec: constraint-satisfiability}.

\begin{definition}
\label{def: WECMDP}
\textbf{Worst-case Escape CMDP} Given a CMDP $M$, the Worst-case Escape CMDP $M_{\mathscr{U}}$ induced on $\mathscr{U}$ is equivalent to the Induced CMDP on $\mathscr{U}$ except that $c(s) = c_{\text{max}}$ for all $s \in \mathscr{U}$.
\end{definition}
Intuitively, the policy that optimises a Worst-case Escape CMDP escapes to the known states before violating the constraint-cost budget under worst-case assumptions.

The following result shows that a policy that is optimised robustly on an estimated Worst-case Escape CMDP will be able to escape the unknown states with high probability while also satisfying the constraint-cost budget within both the (non-estimated) Worst-case Escape CMDP and the full CMDP.
\begin{lemma}
\label{lem: WEL}
\textbf{Worst-case Escape Lemma} Let $M$ be the full CMDP, and $M_{\mathscr{U}}$ be the Worst-case Escape CMDP on $\mathscr{U}$, and $\hat{M}_{\mathscr{U}}$ its estimate. Let $d'$ be the constraint-cost budget within the Worst-case Escape CMDP $M_{\mathscr{U}}$. Let $T' = \inf \{T \in \mathbb{N}: \sum_{t=0}^{T-1} \gamma^{t} c_{\text{max}} \geq d'  \}$, and $T \geq T'$. Moreover, let $\mathcal{P}: \mathcal{S} \times \mathcal{A} \to \Delta^S$ be an uncertainty set, such that for each state-action pair $(s,a) \in \mathcal{S} \times \mathcal{A}$, the true transition dynamics model $P^{*}_{s,a}$ is contained in $\mathcal{P}_{s,a}$ with probability at least $1 - \delta_{\psi}$.\footnote{Note that the proof only applies the uncertainty set on the unknown states, $\mathscr{U}$, since after visiting a state in $\mathscr{K}$, the Worst-case Escape CMDP is terminated. Therefore, for the sake of the lemma, one might equally define the uncertainty set on the unknown states only, i.e. in the form $\mathcal{P}: \mathscr{U} \times \mathcal{A} \to \Delta^S$. However, using the full state space is a convenience data structure in case one desires to also construct an uncertainty set for the known states (see Algorithm~\ref{alg: E4}).} If $\pi$ is a policy in $\hat{M}_{\mathscr{U}}$ that satisfies the constraint of the Worst-case Escape CMDP robustly, that is, 
\begin{align}
\max_{P \in \mathcal{P}} \hat{C}_{P,\pi}(s, T \vert \hat{M}_{\mathscr{U}}) \leq d' \nonumber
\end{align}
for the initial state $s \in \mathscr{U}$, then with probability at least $1 - UA\delta_{\psi}$, $\pi$ in $M_{\mathscr{U}}$ meets the constraint-cost budget $C_{P^{*},\pi}(s \vert M_{\mathscr{U}}) \leq d'$ and takes at most $T'$ steps in $\mathscr{U}$.
\end{lemma}
\textbf{Proof:}\\
Let $\pi$ be a policy in $\hat{M}_{\mathscr{U}}$ that satisfies the constraint on a state $s \in \mathcal{U}$ robustly, i.e. $\max_{P \in \mathcal{P}} \hat{C}_{P,\pi}(s , T \vert \hat{M}_{\mathscr{U}}) \leq d'$. Due to union bound, we have with probability at least $1 - UA\delta_{\psi}$ that $P^{*} \in \mathcal{P}$ (i.e. the true transition dynamics model is contained in the uncertainty set) where $U=\vert\mathscr{U}\vert$. Therefore, with constraint-cost estimates equal to their true values $c_{\text{max}}$ for all steps taken in the worst-case CMDP, we have with probability at least $1 - UA\delta_{\psi}$,
\begin{align}
C_{P^{*},\pi}(s, T \vert M_{\mathscr{U}}) \leq \max_{P \in \mathcal{P}} \hat{C}_{P,\pi}(s, T \vert \hat{M}_{\mathscr{U}}) \leq d'  \,. \nonumber
\end{align}
Since the budget is not exceeded and $T'$ is the infimum of the set $\{T \in \mathbb{N}: \sum_{t=0}^{T-1} \gamma^{t} c_{\text{max}} \geq d'  \}$, the number of steps taken in $\mathscr{U}$ (with $c_t = c_{\text{max}}$) is at most $T'$. 
\qed

\subsection{Balanced wandering}
\label{sec: safe balanced wandering}
To be able to make an unknown state known, each action must be performed often enough, following the Known State Lemma (Lemma~\ref{lem: KS}). In $E^3$, this is achieved through balanced wandering. 
\begin{definition}
\label{def: BW}
For a given state $s \in \mathcal{S}$, \textbf{balanced wandering} takes the action that has been performed the least, 
\begin{equation}
a^{*} = \argmin_{a \in \mathcal{A}} n(s,a) \,.
\end{equation}
\end{definition}

In $E^4$, taking many arbitrary actions with balanced wandering may exceed the constraint-cost budget. How can the agent perform balanced wandering while achieving a safe escape? The answer is to perform balanced wandering as long as the predicted constraint-cost following the escape policy allows doing so, as formalised by the Safe Balanced Wandering Lemma. 
\begin{lemma}
\label{lem: SBW}
\textbf{Safe Balanced Wandering Lemma} Let $M_{\mathscr{U}}$ be the Worst-case Escape CMDP on $\mathscr{U}$, and $\hat{M}_{\mathscr{U}}$ its estimate. Let $d'$ be the constraint-cost budget within the Worst-case Escape CMDP $M_{\mathscr{U}}$. Let $T' = \inf \{T \in \mathbb{N}: \sum_{t=0}^{T-1} \gamma^{t} c_{\text{max}} \geq d'  \}$, and $T \geq T'$. Let $\mathcal{P}: \mathcal{S} \times \mathcal{A} \to \Delta^S$ be an uncertainty set, such that for each state-action pair $(s,a) \in \mathcal{S} \times \mathcal{A}$, the true transition dynamics model $P^{*}_{s,a}$ is contained in $\mathcal{P}_{s,a}$ with probability at least $1 - \delta_{\psi}$.
Let $\pi$ be a policy in $\hat{M}_{\mathscr{U}}$ that satisfies the constraint of the Worst-case CMDP robustly, that is, 
\begin{align}
\max_{P \in \mathcal{P}} \hat{C}_{P,\pi}(s, T \vert \hat{M}_{\mathscr{U}}) \leq d' \nonumber
\end{align}
for the initial state $s \in \mathscr{U}$. Let $\mu$ be a strategy that performs balanced wandering but switches to policy $\pi$ at time $t$ when
\begin{align}
C(p_t) + \gamma^{t} \max_{P \in \mathcal{P}} \max_{a_t \in \mathcal{A}} \sum_{s_{t+1} \in \mathcal{S}} P_{s_t,a_t}(s_{t+1}) \hat{C}_{P,\pi}(s_{t+1}, T \vert \hat{M}_{\mathscr{U}}) \geq d' - c_{\text{max}} \,, \nonumber
\end{align}
where $p_t=\{s,s_1,\dots,s_{t-1}\}$ is the $t$-step path taken so far from $s$. Then with probability at least $1 - UA\delta_{\psi}$, $\mu$ is safe, in the sense that 
\begin{align}
C_{P^{*},\mu}(s, T \vert M_{\mathscr{U}}) \leq  d' \,. \nonumber
\end{align}
\end{lemma}
\textbf{Proof:}\\
Let $t \in \{0,\dots,T'-1\}$ be the first step for which
\begin{align}
C(p_t) +  \gamma^t \max_{P \in \mathcal{P}} \max_{a_t \in \mathcal{A}} \sum_{s_{t} \in \mathcal{S}} P_{s_t,a_t}(s_{t+1}) \hat{C}_{P,\pi}(s_{t+1}, T \vert \hat{M}_{\mathscr{U}}) \geq d' - c_{\text{max}} \,. \nonumber
\end{align}
Since one step expends at most $c_{\text{max}}$, this implies 
\begin{align}
C(p_t) + &\gamma^t \max_{P \in \mathcal{P}} \hat{C}_{P,\pi}(s, T \vert \hat{M}_{\mathscr{U}}) \leq C(p_t) + \nonumber \\
 & \quad \gamma^t \max_{P \in \mathcal{P}} \max_{a_t \in \mathcal{A}} \sum_{s_{t+1} \in \mathcal{S}} P_{s_t,a_t}(s_{t+1}) \hat{C}_{P,\pi}(s_{t+1}, T \vert 
 \hat{M}_{\mathscr{U}}) \leq d' \nonumber
\end{align}
and therefore, by Lemma \ref{lem: WEL}, with probability $1 - UA\delta_{\psi}$ that
\begin{align}
C(p_t) + \gamma^t C_{P^{*},\mu}(s_t, T \vert M_{\mathscr{U}}) \leq d' \,. \nonumber
\end{align}
Since this holds for any path $p_t$ from $s$, it also holds on expectation over $t$-step paths from $s$, such that with probability $1 - UA\delta_{\psi}$
\begin{align}
C_{P^{*},\mu}(s, T \vert M_{\mathscr{U}}) \leq d' \,. \nonumber
\end{align}
\qed

\subsection{Escape budget lemma}
\label{sec: EB-lemma}
Having introduced how to use induced CMDPs for safety within known and unknown states, this section now turns to the full CMDP $M$: that is, when following T-step policies on induced CMDPs $M_{\mathscr{K}}$ and $M_{\mathscr{U}}$, does this still guarantee safety across the entire CMDP? To resolve this question, the following lemma, called the Escape Budget Lemma, formulates an ``\textbf{escape budget}'' $d' \leq d$, the highest constraint-cost in $M_{\mathscr{U}}$ that still provides a probabilistic safety guarantee over the entire CMDP $M$.

First follows a definition of safe return states, which indicate states from which a $T$-step trajectory is known to stay in $\mathscr{K}$ following some policy $\pi$ and which satisifes a much lower budget but has no requirements on the value (so typically yields a poor value).
\begin{definition}
\label{def: safe return}
A \textbf{$d_s$-safe return state} is a known state  $s \in \mathscr{K}$ for which there exists a stationary policy $\pi_{d_s}: \mathcal{S} \to \mathcal{A}$ that takes at least $T$ steps from $s$ in $\mathscr{K}$ with probability $1 - \delta$ and that satisfies $C_{\pi_{d_s}}(s,T \vert \gamma=1) \leq d_s - \epsilon$. This policy $\pi$ is called a $d_s$-safe return policy and the set of such states is denoted by $\mathscr{K}_{d_s}$. 
\end{definition}
While the CMDP of interest, $M$ , has $\gamma \in [0,1)$, the above definition is
based on $\gamma=1$ to ensure accounting for the worst-case impact of safe return
on future constraint-satisfaction. Since $T$ is the $\epsilon$-horizon time, this implies any non-stationary policy $\pi_n$ which first applies $\pi_{d_s}$ for $T$ steps and then any policy in $M$ will yield $C_{\pi_n}(s) \leq d_s$ for any starting state $s \in \mathscr{K}_{d_s}$.

The Escape Budget Lemma takes place in a setting where there is an cycle alternating between three policies:
\begin{itemize}
\item an exploration policy in $M_{\mathscr{K}}$, which aims to find an unknown state;
\item a safe balanced wandering policy in $M_{\mathscr{U}}$, which performs balanced wandering followed by an escape back to the known states, in an attempt to make states known; and
\item a safe return policy in $M_{\mathscr{K}}$, which acts $T$ steps within $\mathscr{K}$ with low expected $T$-step constraint-cost, to ensure long-term constraint-satisfaction despite the constraint-costs of exploration attempts.
\end{itemize}
This alternation defines a non-stationary policy $\pi = \{\pi_{i}\}_{i=0}^{\infty}$, where $i$ indicates the particular policy being used, where $i=0,3,6,\dots$ index exploration policies, $i=1,4,7,\dots$ index safe balanced wandering policies, and $i=2,5,8,\dots$ index safe return policies. Within this setting, the lemma determines the available escape budget based on the initial path cost in the known states, the worst-case expected cost in the unknown states ($d'$), and the worst-case expected asymptotic constraint-cost of the safe return policy ($d''$).
\begin{lemma}
\label{lem: EB}
\textbf{Escape Budget Lemma} Let $M$ be the full CMDP, $M_{\mathscr{K}}$ be the known-state CMDP, $M_{\mathscr{U}}$ be the Worst-case Escape CMDP, and let $\mathscr{K}_{d_s} = \mathscr{K}$ for some level $d_s \leq d$. Let $p$ be a path from $s \in \mathscr{K}$ to a starting state $s' \in \mathscr{U}$. Let $d'' \leq d - \epsilon$ be the budget in $M_{\mathscr{K}}$ and $d'_i$ be the ``escape'' budget in $M_{\mathscr{U}}$ at the $i$'th visitation to $M_{\mathscr{U}}$. Let $\pi = \{\pi_{i}\}_{i=0}^{\infty}$ be a non-stationary policy, defined on $M_{\mathscr{K}}$ with expected asymptotic cost $C_{\pi_i}(s \vert M_{\mathscr{K}}) \leq d''$ for $i=0,3,6,\dots$, on $M_{\mathscr{U}}$ with expected asymptotic cost $C_{\pi_i}(s \vert M_{\mathscr{K}}) \leq d'_{i//3}$\footnote{The notation $//$ indicates integer division} for $i=1,4,7,\dots$, and on $M_{\mathscr{K}}$ as a $d_s$-safe return policy for $i=2,5,8,\dots$. Further, let $\Pi(s,d'') \neq \emptyset$. Let $d_{\text{min}}' = d - d'' - 2d_s + \epsilon$ and $T_{\text{min}} = \inf \{T \in \mathbb{N}: \sum_{t=0}^{T-1} \gamma^{t} c_{\text{max}} \geq d_{\text{min}}' \}$. Finally, define $D^{+}:=D(M)+1$, where $D(M)$ is the diameter of $M$. Define the following requirements:\\
\textbf{(a) Diameter requirement:} the diameter satisfies $D^{+} \leq T_{\text{min}} \leq T'_{i//3}$; \\
\textbf{(b) Known-state requirements:}  for any recent path $p$ from $M_{\mathscr{K}}$ to unknown state $s_0 \in \mathscr{U}$, require $d'_i \leq \frac{1}{\gamma^{T_k}}  \left( d - \gamma^{T_k+1} d_s  -  C(p_k) \right)$; \\
\textbf{(c) Unknown-state requirement:} $d'_i \leq d - \gamma d_s$; and\\
\textbf{(d) Safe return requirement:} $d_s \leq \frac{1}{2}\left(d + \epsilon - d'' - D^{+} c_{\text{max}}\right)$.\\
Then these requirements are not conflicting and imply $C_{\pi}(s_t) \leq d$ for all $s_t \in \mathcal{S}$ at all times $t$. Moreover, the budget allows at least one time step of balanced wandering before escape.
\end{lemma}
\textbf{Proof:}\\
The proof will first show that the requirements (a), (b), (c), and (d) are not conflicting with each other. Then, the proof selects an arbitrary known state $s_t$ at any time $t$ and shows how the requirements yield $C_{\pi}(s_t) \leq d$. Then, the proof analogously selects an unknown state $s_t$ at any time $t$ and shows how the requirements yield $C_{\pi}(s_t) \leq d$. Finally, the proof shows that at least one time step of balanced wandering is allowed due to the definition of $D^+$.\\
\textbf{0) Requirements are not conlicting.} Requirements \textbf{(b)} and \textbf{(c)} are not conflicting with each other: since they are both upper bounds, one can set $d'_{i//3} =\min\left(\frac{1}{\gamma^{T_k}}  \left( d - \gamma^{T_k+T'_{i//3}} d_s  -  C(p_k) \right), d - \gamma^{T'_{i//3}} d''  \right)$. Now it is required to show that \textbf{(a)} does not conflict with this setting of $d'_{i//3}$. First, note that the worst case, where the discount factor is $\gamma=1$ and where $d_s - \epsilon + d''$ is expended in the known states and $d_s$ is expended from a safe return state, yields the lower bound for $d'_{i//3}$,
\begin{align} 
d - d'' + \epsilon - 2d_s  = d_{\text{min}} \leq  d'_{i//3} \,, \nonumber
\end{align}
and therefore $T_{\text{min}} \leq d'_{i//3}$.
Filling in $d_s$ according to requirement \textbf{(d)} yields 
\begin{align} 
d'_{i//3} &\geq d - d'' + \epsilon - 2d_s \nonumber \\ 
		  &\geq d - d'' + \epsilon - 2( \frac{1}{2}(d  - d'' + \epsilon -D^{+}c_{\text{max}})) = D^{+}c_{\text{max}} \,, \nonumber
\end{align}
and therefore
\begin{align} 
T'_{i//3} \geq d'_{i//3}/c_{\text{max}} \geq D^{+} \,, \nonumber
\end{align}
which is consistent with requirement \textbf{(a)}.\\
\textbf{1) Known states.} Let $s \in \mathscr{K}$. If the agent is at the starting state of its $d_s$-safe return then the proof is finished by definition of $d_s\leq d$. Otherwise, the agent is either somewhere along the trajectory of the $d_s$-safe return policy or it is along the trajectory of the exploration policy. In both cases, the agent forms a path $p_k$ from $s$ to a starting unknown state $s_u \in \mathscr{U}$, then follows a safe balanced wandering policy that escapes to a $d_s$-safe return state $s_k$ with probability $1 - UA\delta_{\psi}$, and then from $s_k$ takes $T$ steps using a $d_s$-safe return policy. Denoting the length of the path $p_k$ as $T_k$, the cost of the path in the known states as $C(p_k)$, the cost of the path in the unknown states as $C(p_u)$, and the expected asymptotic constraint-cost from the safe return states as $C_{\pi}(s_k)$,  the statement to prove is
\begin{align}
C_{\pi}(s) =  \mathbb{E}_{P^*,\pi} \left [ C(p_k) + \gamma^{T_k} C(p_u) + \gamma^{T_k+T_u} C_{\pi}(s_k )  \right] \leq d \,. \nonumber 
\end{align}
By definition of the diameter, requirement \textbf{(a)} ensures that there exists an escape policy $\pi'$ which satisfies $C_{\pi'}(s \vert M_{\mathscr{U}}) \leq d'_{i//3}$ from any $s \in \mathscr{U}$. Therefore, setting $\pi_i=\mu$, where $\mu$ is a safe balanced wandering policy for the budget $d'_{i//3}$, we have $\mathbb{E}\left[C(p_u)\right] \leq d'_{i//3}$ and $T_u \geq 1$. Moreover, the safe return policy ensures that $C_{\pi}(s) \leq d_s$ such that
\begin{align}
&C(p_k) + \gamma^{T_k} \mathbb{E}\left[C(p_u) + \gamma^{T_u} C_{\pi}(s_k) \right] \nonumber \\
&\leq C(p_k) +  \gamma^{T_k} d'_{i//3} + \gamma^{T_k+1} d_s \nonumber \\
&\leq d \,,  \nonumber 
\end{align}
where the last line follows from requirement \textbf{(b)}.
Since the inequality holds for any path $p_k$, it also holds on expectation over paths.\\
\textbf{2) Unknown states.} Let $s \in \mathscr{U}$. The agent forms a path $p_u$ from $s$ to a starting unknown state $s_k \in \mathscr{K}$. Denoting the length of the path as $T_u$, based on the cost of the path $C(p_u)$ and the expected asymptotic cost over the terminal state in $\mathscr{K}$, the statement to prove is
\begin{align*}
C_{\pi}(s) &=  \mathbb{E}_{P^*,\pi} \left [ C(p_u) + \gamma^{T_u} C_{\pi}(s_k) \right] \leq d \,. \nonumber \\
\end{align*}
Note that $T_u \geq 1$, and via requirement \textbf{(a)} and \textbf{(b)}, $C_{\pi}(s \vert M_{\mathscr{U}}) = C_{\pi_i}(s \vert M_{\mathscr{U}}) \leq d'_{i//3}$ for some stationary policy $\pi_i$,  and $C_{\pi}(s_k) \leq d_s$. Further filling in requirement \textbf{(c)} yields the statement to prove,
\begin{align}
C_{\pi}(s) &\leq d'_{i//3} + \gamma d_s \nonumber \\ 
		   &\leq d - \gamma d_s + \gamma d_s = d \,. \nonumber
\end{align}
\textbf{3) At least one step of balanced wandering.}  Let $j \in \{0,1,\dots\}$. As shown in part \textbf{0)} of the proof, $T'_{j} \geq D^{+} = D(M) + 1$. By definition of the diameter, an escape policy can return to the known states within at most $D(M)$ time steps. Therefore, at least one step of balanced wandering can be performed without exceeding the budget $d'_j$.
\qed

For the $C_{\pi}(s_t) \leq d$ for all $t$ in the current trajectory, it is required to perform computation b) over all known states recently visited and take the minimum, and then again the minimum between the result and the quantity in c). For such a computation, one can take the last $T$ steps within $\mathscr{K}$ and subtract $\epsilon$ for the computation of $d'$. Note that for unknown states, one only needs to perform c) on starting states because it assumes the worst-case $c_{\text{max}}$ for all coming $T'$ steps whereas intermediate unknown states will only yield $c_{\text{max}}$ for a lower number of steps.

The Escape Budget Lemma hereby provides a general strategy for determining a desired escape budget. Below is an example of a high constraint-cost trajectory in $\mathscr{K}$ and the resulting budget $d'$. The below trajectory is shorter than usual purely for demonstration purposes; any trajectory (except the initial $i=0$ trajectory) would involve at least $T$ steps from the known states due to the $d_s$-safe policy taking $T$ steps prior to the exploration policy.
\begin{example} \label{ex: 1} Let $c_{\text{max}}=1$, $\gamma=0.98$, and $\epsilon=1.0$. The setting yields an $\epsilon$-horizon time $T=\lceil \frac{1}{1 - \gamma} \ln\left(\frac{c_{\text{max}}}{\epsilon(1-\gamma)}\right) \rceil=196$. Further, $d=25$, and safe return policies exist for all $s \in \mathscr{K}$ with $d_s=5.0$, both of which are well below the general upper bound $G_{\text{max}}^c(T) + \epsilon=49.05$. The diameter of the full CMDP is known to be $D(M)=3$ and therefore set $D^{+}=4$. Compute the known-state budget as $d'' = d - 2d_s - D^{+} + \epsilon = 12.0$. Computing $T_{\text{min}} =\inf \{T \in \mathbb{N}: \sum_{t=0}^{T-1} \gamma^{t} c_{\text{max}} \geq d - d'' - 2d_s + \epsilon \} = 3$, $D^{+} = T_{\text{min}}$ implies the constraint on the unknown states is certainly satisfiable for any trajectory. The current trajectory of constraint-cost yields $[0, c_{\text{max}}, 0 , 0, c_{\text{max}},c_{\text{max}}/2,c_{\text{max}},c_{\text{max}},c_{\text{max}}]$, resulting in a list of path-based constraint-costs to check, $C_p = [3.29, 3.66, 2.95, 3.28, 3.65, 2.94, 2.71, 1.9, 1.0]$ for $t=0, \dots,8$. For condition (b), looping over $C \in C_p$ and computing $d' \gets \frac{1}{\gamma^{T_k}} \left( d - \gamma^{T_k + 1}d_s -  C(p) \right)$, $d'$ is minimal for $t=4$, namely $d'=17.53$. Condition (c) does not require any adjustment to $d'$. This setting of $d'$ leaves $T'=21$ steps from which at least $T' - D(M) = 18$ steps of balanced wandering can be allocated. Following escape, the safe return policy will take $T$ steps in the known states, where it yields $C_{\pi}(s,T \vert \gamma=1) \leq d_s - \epsilon = 4.0$, before an exploration policy starts an exploration attempt.
\end{example}

\subsection{Simulated Budget Satisfaction Lemma}
\label{sec: SBS-lemma}
Previously, the $l$-safe Explore-or-Exploit Lemma (Lemma~\ref{lem: lEE}) showed that, if a policy yields $d - 2\epsilon$-safety for the expected $T$-step constraint-cost in a simulation of the Known-state CMDP, then it also yields $d$-safety for the expected asymptotic constraint-cost in the real Known-state CMDP. The converse does not follow automatically; if a policy $\pi$ yields $d$-safety for the expected asymptotic constraint-cost  in a real Known-state CMDP, then $d-2\epsilon$-safety is not guaranteed for the $T$-step expected constraint-cost in a simulation of the Known-state CMDP -- making it possible that the constraint cannot be satisfied. The following lemma provides the conditions on the real-world CMDP under which the $d - 2\epsilon$-safety constraint can be satisfied in simulation.
\begin{lemma}
\label{lem: SBS}
\textbf{Simulated Budget Satisfaction Lemma} Let $M$ be a full CMDP, $M_{\mathscr{K}}$ be a known-state CMDP induced over a set of known states $\mathscr{K}$ in $M$, $s \in \mathscr{K}$, $\pi$ be a policy that satisfies $\pi \in \Pi_c(s,l - \epsilon)$ over $M$, where $l = d-2\epsilon$. Further, let $\hat{M}_{\mathscr{K}}$ be an estimation of $M_{\mathscr{K}}$. Then
$\pi$ will satisfy $\hat{C}_{\pi}(s,T \vert \hat{M}_{\mathscr{K}}) \leq l$.
\end{lemma}
\textbf{Proof:}\\
Note that $C_{\pi}(s \vert M_{\mathscr{K}}) \leq C_{\pi}(s)$ due to constraint-costs being positive. Therefore, by the Constrained T-step Estimation Lemma (Lemma~\ref{lem: CTE}) and the Constrained Simulation Lemma (Lemma~\ref{lem: CS}), 
\begin{align}
\hat{C}_{\pi}(s, T \vert \hat{M}_{\mathscr{K}}) \leq \hat{C}_{\pi}(s \vert \hat{M}_{\mathscr{K}}) \leq C_{\pi}(s \vert M_{\mathscr{K}}) + \epsilon \leq C_{\pi}(s) + \epsilon \leq l - \epsilon + \epsilon = l \,. 
\nonumber 
\end{align}
Thereby, $\hat{C}_{\pi}(s, T \vert \hat{M}_{\mathscr{K}}) \leq l$. \qed

\subsection{Proof of Theorem 1}
\label{sec: together}
With all the lemmas in place, the following presents the step-by-step proof of Theorem 1.

\paragraph{(a) Returning a near-optimal constrained policy within polynomial time.} First, we prove the statement that for $\epsilon \geq 0$, $E^4$ outputs with probability at least $1 - \delta$ a near-optimal constrained policy $\pi$ for $s$ with $V_{\pi}(s) \geq V_{\pi^{*}}(s) - \epsilon$ and $C_{\pi}(s) \leq d$ with sample complexity and time complexity that is polynomial in $1/\epsilon$, $1/\delta$, $S$, the horizon time $1/(1-\gamma)$, $r_{\text{max}}$, and $c_{\text{max}}$.

Select $s \in \mathcal{S}$ and $\epsilon\geq 0$ arbitrarily. Let $m_{\text{known}}$ be $O((STG/\epsilon)^4 \text{Var}_{\text{max}} \ln(1/\delta'))$, where $\delta' = \delta/4$. Define $T$ as the $\epsilon/4$-horizon time. If $s \in \mathscr{K}$, then by assumption $\Pi_c(s,d,T) = 
\{\pi \in \Pi: C_{\pi}(s,T) \leq d   \} \neq \emptyset$. By Lemma~\ref{lem: CEE}, either there exists a near-optimal constraint-satisfying policy $\pi \in \Pi_c(s,d,T)$ from $s$ with $V_{\pi}(s,T \vert M_{\mathscr{K}}) \geq V_{\pi^*}(s,T) - \epsilon/2$ or there exists a policy $\pi' \in \Pi_c(s,d,T)$ that finds an unknown state with probability at least $\epsilon/(2G_{\text{max}}^r(T))$, regardless of the choice of $T$ and $\epsilon$. If $\hat{V}_{\pi}(s,T \vert \hat{M}_{\mathscr{K}}) \geq V_{\pi^*}(s) - \epsilon/2$, then due to Lemma~\ref{lem: IU}, Lemma~\ref{lem: CTE}, and Lemma~\ref{lem: CS}, it follows with probability $1 - \delta'$ that
\begin{align}
V_{\pi^*}(s) - \epsilon/2 \leq \hat{V}_{\pi}(s,T \vert \hat{M}_{\mathscr{K}}) \leq \hat{V}_{\pi}(s,T) \leq \hat{V}_{\pi}(s) \leq V_{\pi}(s) + \epsilon/4  \nonumber
\end{align}
and that 
\begin{align}
V_{\pi}(s,T \vert M_{\mathscr{K}}) \geq \hat{V}_{\pi}(s,T \vert \hat{M}_{\mathscr{K}}) - \epsilon/4 \geq V_{\pi^*}(s) - 3\epsilon/4 \geq V_{\pi^*}(s,T) - 3\epsilon/4 \,, \nonumber
\end{align}
which implies offline optimisation will find an $\epsilon$-optimal exploitation policy $\pi$, and the proof is finished in this case.
 Otherwise, if $\hat{V}_{\pi'}(s,T \vert \hat{M}_{\mathscr{K}}) < V_{\pi^*}(s) - \epsilon/2$ this implies there exists an exploration policy that can be found by $E^4$ since 
\begin{align}
V_{\pi'}(s,T\vert M_{\mathscr{K}}) \leq \hat{V}_{\pi'}(s,T \vert \hat{M}_{\mathscr{K}}) + \epsilon/4 < V_{\pi^*}(s) - 3\epsilon/4 \leq V_{\pi^*}(s,T) - \epsilon/2 \,. \nonumber
\end{align}
$E^4$ starts an exploration attempt using $\pi'$, which takes a $T$-step trajectory to find an unknown state, based on the exploration known-state CMDP. Such attempts may fail repeatedly but have a success probability of at least $\epsilon/(2G_{\text{max}}^r(T))$. Upon success, the algorithm performs balanced wandering in $M_{\mathscr{U}}$. This cycle of attempted explorations is repeated and due to the repeated visitation, states will become known after a number of visitations $m_{\text{known}} = O((STG/\epsilon)^4 \text{Var}_{\text{max}} \log(1/\delta'))$. In the worst case, the algorithm must make all the states known before finding a near-optimal constrained exploitation policy for $s$, requiring up to $S m_{\text{known}}$ steps of balanced wandering. In the worst case, due to $T' \geq D(M) + 1$ following the requirements of Lemma~\ref{lem: EB}, there is only one step of balanced wandering per successful exploration attempt. With this in mind, Chernoff bound analysis (see Appendix C) shows that with probability $1 - \delta'$, the total number of exploration attempts before making all states known is bounded by $O\left(\frac{G_{\text{max}}^r(T)}{\epsilon} Sm_{\text{known}} \ln(S/\delta') \right)$. Since each $T$-step trajectory takes at most $T$ actions, the number of actions taken by $E^4$ before a near-optimal policy can be returned will be bounded by 
\begin{equation}
O\left(T \frac{G_{\text{max}}^r(T)}{\epsilon} Sm_{\text{known}} \log(S/\delta') \right) \,.
\end{equation}
Therefore, with $T \geq \ln\left(\frac{\max(r_{\text{max}},c_{\text{max}})}{\epsilon(1-\gamma)}\right) \frac{1}{1-\gamma}$ and $m_{\text{known}} = O((STG/\epsilon)^4 \text{Var}_{\text{max}} \log(1/\delta'))$, the sample complexity to output a near-optimal constrained policy for state $s \in \mathcal{S}$ is polynomial in $1/\epsilon$, $1/\delta$, $S$, the horizon time $1/(1-\gamma)$, $r_{\text{max}}$ and $c_{\text{max}}$.\footnote{The dependencies on $r_{\text{max}}$ and $c_{\text{max}}$ follow from the factors $G_{\text{max}}^r(T)$, $G^4$, and $\text{Var}_{\text{max}} \leq \max(c_{\text{max}},r_{\text{max}})^2 / 4$.} Since offline optimisation is repeated at every attempted exploration, the time complexity of $E^4$ is 
\begin{equation}
O\left(\texttt{Opt} \, T \frac{G_{\text{max}}^r(T)}{\epsilon} Sm_{\text{known}} \log(S/\delta') \right) \,, 
\end{equation}
where \texttt{Opt} refers to the time complexity of the offline optimisation. Given the above, all that remains to be shown is that 1) \texttt{Opt} is polynomial-time; and 2) offline optimisation converges to the global optimum. These statements are demonstrated for different offline optimisation algorithms in Section~\ref{sec: offline}. Summing the three observed failure probabilities, the above results hold with probability at least $1 - \delta$. \qed

\paragraph{(b) Safe exploration.} Here, we prove the statement that at any time $t$, the non-stationary policy $\pi^n=\{\pi_{i}\}_{i=1}^{\infty}$ induced by the exploration process of $E^4$ yields $C_{\pi^n}(s_t) \leq d$, as specified in the CMDP objective, with probability at least $1 - (UA\delta_{\psi} + \delta)$.

Select an arbitrary time point $t$ in the RL agent's lifetime and observe the state $s_t$, let $\epsilon > 0$, and let $d''$ be the known-state budget, and let $l = d'' - 2\epsilon$. Moreover, it is assumed that $\Pi_c(s_t,l - \epsilon,T) = 
\{\pi \in \Pi: C_{\pi}(s_t,T) \leq  l - \epsilon  \} \neq \emptyset$, $\Pi_c(s_t,d_s - \epsilon,T) = 
\{\pi \in \Pi: C_{\pi}(s_t,T) \leq  d_s - \epsilon  \} \neq \emptyset$ and that the optimal value within $\Pi_c(s_t,l-\epsilon,T)$ is $\epsilon$-close to that of the optimal value within $\Pi_c(s_t,d,T)$.\footnote{Note that the last assumption is not needed for safety throughout exploration; instead its purpose is to be able to return an $\epsilon$-optimal exploitation policy despite the difference between $d''$ and $d$. If the assumption does not hold on the initial known-state budget $d''$, one potential way to realise the assumption practically is to reset $d''$ to be close to $d$ after most states have been made known, thereby returning the exploitation policy only when the known-state budget is sufficiently close to $d$.}

If $s_t \in \mathscr{K}$, then either the agent is performing $d_s$-safe return or the agent is doing exploration/exploitation. If the agent is at the starting state of its $d_s$-safe return then the proof is finished by definition of $d_s$ (see Definition~\ref{def: safe return}). If not, the exploration or exploitation policy will be activated after fewer than $T$ steps of $d_s$-safe return. The setting $\Pi_c(s_t,l - \epsilon,T) =  \emptyset$ implies that the offline optimisation constraints are satisfiable on level $l$ in $\hat{M}_{\mathscr{K}}$ with probability $1 - \delta'$. By Lemma~\ref{lem: lEE}, any policy $\pi$ with $\hat{C}_{\pi}(s_t, T\vert \hat{M}_{\mathscr{K}}) \leq l$ yields $C_{\pi}(s_t \vert M_{\mathscr{K}}) \leq d''$ as required. By Lemma~\ref{lem: CEE} either the exploitation policy $\pi' \in \Pi_c(s_t,d'',T)$ exists or the exploration policy $\pi'' \in \Pi_c(s_t,d'',T)$ exists (or both). By assumption, the optimal value within $\Pi_c(s_t,d,T)$ is $\epsilon$-close to that of the optimal value within $\Pi_c(s_t,l-\epsilon,T)$ and consequently that in $\Pi_c(s_t,d'',T)$ as well. Combining the above, either a) $E^4$ will be able to find a policy $\pi' \in \Pi_c(s_t,d'',T)$ that is near-optimal from $s_t$, or b) $E^4$ finds a policy $\pi''$ which has $C_{\pi''}(s_t,T \vert M_{\mathscr{K}})\leq d''$ and which with probability $p > \epsilon/G^r_{\text{max}}(T)$ reaches un unknown state in $T$ steps.  If the agent stays within $\mathscr{K}$, then the proof for the known states is complete since then $C_{\pi}(s_t) = C_{\pi}(s_t \vert M_{\mathscr{K}}) \leq d '' \leq d$. Otherwise, by Lemma~\ref{lem: EB}, the  setting of $d'$ implies that the non-stationary policy $\pi^{n}$, which alternates between $l$-safe exploration/exploration, $d'$-safe balanced wandering, and $d_s$-safe return, yields $C_{\pi^{n}}(s_t) \leq d$ -- provided that the agent does not exceed the budget $d'$ in the unknown states (see following paragraph).

If $s_t \in \mathscr{U}$, then the $E^4$ agent will use the safe balanced wandering policy $\mu$ until returning to the known states. Let $\mathcal{P}$ be an uncertainty set such that 1) for all $(s,a) \in \mathcal{S} \times \mathcal{A}$, $\mathcal{P}_{s,a}$ contains $P_{s,a}^{*}$ with probability at least $1 - \delta_{\psi}$, and 2) its worst-transition diameter is at most $T' - 1$ (see Eq.~\ref{eq: worst-transition}). From the definition of $T' = \inf \{T \in \mathbb{N}: \sum_{t=0}^{T-1} \gamma^{t} c_{\text{max}} \geq d'  \}$, there exists a policy $\pi''$ (the escape policy) that satisfies
\begin{align}
\max_{P \in \mathcal{P}} \hat{C}_{P,\pi''}(s, T \vert \hat{M}_{\mathscr{U}}) \leq d' \,. \nonumber
\end{align}
By Lemma~\ref{lem: WEL}, the above implies $C_{P^{*},\pi''}(s, T \vert M_{\mathscr{U}}) \leq \max_{P \in \mathcal{P}} \hat{C}_{P,\pi''}(s, T \vert \hat{M}_{\mathscr{U}}) \leq d'$  with  probability at least $1 - UA\delta_{\psi}$. Now consider the trajectory within the unknown states. $\mu$ first performs steps of balanced wandering until taking one more step would not allow the escape policy $\pi''$ to return safely, i.e. as soon as $C(p_t) + \gamma^{t} \max_{P \in \mathcal{P}} \max_{a_t \in \mathcal{A}} \sum_{s_{t+1} \in \mathcal{S}} P_{s_t,a_t}(s_{t+1}) \hat{C}_{P,\pi''}(s_{t+1}, T \vert \hat{M}_{\mathscr{U}}) \geq d' - c_{\text{max}}$. By the proof of Lemma~\ref{lem: SBW}, the total cost within the unknown states is
\begin{align}
c(s_t) + \gamma^t C_{P^{*},\pi}(s, T \vert M_{\mathscr{U}}) &\leq c(s_t) + \gamma^t C_{\mu}(s_t, T \vert M_{\mathscr{U}}) \nonumber \\ 
&\leq C(p_t) + \gamma^t C_{\mu}(s_t, T \vert M_{\mathscr{U}}) \leq d' \nonumber
\end{align}
with probability at least $1 - UA\delta_{\psi}$. 

This constraint-satisfaction implies that $C_{\pi^{n}}(s_t) \leq d$ when exploring or exploiting from $s_t \in \mathscr{K}$. Similarly, due to requirement (c) in Lemma~\ref{lem: EB}, it follows that for $s_t \in \mathscr{U}$, $C_{\pi^n}(s_t) \leq d$. Further, the safe return policy may fail to stay within $\mathscr{K}$ with probability $\delta'$. Therefore, combining all the previous yields $C_{\pi^n}(s_t) \leq d $ with probability at least $1 - (UA\delta_{\psi} + \delta)$.

\qed

\section{Explicit Explore, Exploit, or Escape algorithm}
\label{sec: E4}
With the above theory in mind, $E^4$ is now further developed as a practical algorithmic framework, with a discussion of different prototype implementations. 
Within the main loop of the algorithm, three different policies are optimised: an exploration policy, an exploitation policy, and an escape policy. We discuss a range of offline methods to optimise these policies, including policy gradient, linear programming and dynamic programming approaches (Section~\ref{sec: offline}). The general flow of how these policies fit together is then discussed in Section~\ref{sec: EEEpolicies}, with the complete algorithm being summarised in Algorithm~\ref{alg: E4}. The algorithm and theory relies on two conditions for robust constraint-satisfiability, namely that the diameter of the CMDP must be limited and the availability of tight uncertainty sets that still capture the true transition dynamics with high probability. These conditions are discussed along with a variety of example uncertainty sets (Section ~\ref{sec: constraint-satisfiability}).
\subsection{Offline optimisation}
\label{sec: offline}
The algorithm starts by performing offline optimisation of three different policies, the exploration policy, the exploitation policy, and the escape policy, each of which  are defined over a different induced CMDP. The remainder of this subsection provides four possible approaches for offline optimisation in $E^4$, comparing their time complexity, applicability, and scalability. Using one of these algorithms allows fulfilling the two remaining conditions on  offline optimisation in Section \ref{sec: theory}, namely 1) the algorithm is polynomial-time; and 2) the algorithm converges to the global optimum. Therefore, using these algorithms for $E^4$ yields near-optimal constrained policies in polynomial time.
\subsubsection{Robust-constrained policy gradient}
A first proposed approach to optimising these policies is based on an existing solution, namely the RCMDP policy gradient \citep{Russel2020}, where distributional robustness is integrated into CMDPs. This is proposed for the Worst-case Escape CMDP but can optionally also be used for the Known-state CMDPs (see Algorithm \ref{alg: E4}). As \cite{Russel2020} note, one can incorporate the argmax over either the value function or the constraint-cost and they choose the value function; in $E^4$, safety is the main concern, and therefore robustness is incorporated by an argmax over the constraint-cost.
The robust CMDP objective is solved by computing the saddle point of the Lagrangian for a given budget $d$,
\begin{equation}
\label{eq: Lagrange}
\min_{\lambda \geq 0} \max_{\pi_{\theta}} L(\lambda,\pi_{\theta}) =  \hat{V}_{\pi_{\theta},P}(s, T \vert \hat{M}_{\mathscr{S}}) - \lambda \left(d -  \hat{C}_{\pi_{\theta},P}(s, T\vert \hat{M}_{\mathscr{S}})   \right) \,,
\end{equation}
where $\hat{M}_{\mathscr{S}}$ is an induced CMDP over $\mathscr{S} \subset \mathcal{S}$, $\mathcal{P}$ is the uncertainty set, and $P=\argmax_{P' \in \mathcal{P}} \hat{C}_{\pi_{\theta},P'}(s, T \vert \hat{M}_{\mathscr{S}})$. Substituting $f(\theta) = \hat{V}_{\pi_{\theta},P}(s \vert \hat{M}_{\mathscr{S}})$ and $g(\theta) = d - \hat{C}_{\pi_{\theta},P}(s, T\vert \hat{M}_{\mathscr{S}}) $, the aim is to find the policy $\pi_{\theta}$ such that the gradient is a null-vector; that is,
\begin{equation}
\nabla_{\theta} L(\lambda,\pi_{\theta}) = \frac{df(\theta)}{d\theta} - \lambda \frac{dg(\theta)}{d\theta} = \mathbf{0} 
\end{equation}
and
\begin{equation}
\nabla_{\lambda} L(\lambda,\pi_{\theta}) = g(\theta) = \mathbf{0} \,.
\end{equation}
To optimise the above objective, sampling of limited-step trajectories is repeated for a large number of independent iterations starting from a randomly selected state in the subset of the state space over which the CMDP is induced (see Algorithm \ref{alg: optimisation}). Based on the large number of trajectories collected, one then performs gradient descent in $\lambda$ and gradient ascent in $\theta$.  A benefit of this approach is scalability to a high number of parameters.

A common wisdom is that gradient descent methods do not perform well in multi-modal and non-convex landscapes, where they may get stuck in local optima. Similarly, the theory behind the RCMDP policy gradient only provides a convergence result for a local optimum \citep{Russel2021,Russel2020}. However, with sufficient over-parametrisation -- and more precisely, with a number of neurons $n_{h}$ polynomial in the number of samples $n_{\xi}$, number of hidden layers $l$, and the (inverse of) the maximal distance between data points $\nu^{-1}$ -- a neural network can learn the global optimum to arbitrary accuracy within polynomial number of samples \citep{Allen-Zhu2019}. In addition to general results for the $L2$-norm, \cite{Allen-Zhu2019} provides similar results for arbitrary Lipschitz-continuous loss functions; one such result states that if the loss function is $\sigma$-gradient dominant (see \cite{Zhou2017} for its definition), then gradient descent finds with probability $1 - e^{-\Omega(\ln^2(n_{\xi}))}$ an $\epsilon$-optimal parameter vector within a number of iterations $I$ polynomial in $n_h$, $l$, $\nu^{-1}$, $\sigma^{-1}$, and $\epsilon^{-1}$. Per iteration, all operations have polynomial time complexity: look-ups, list appending, random number generation, gradient computation, division, and the forward pass of a neural network. Therefore, the resulting algorithm has overall polynomial time complexity and converges with high probability to the global optimum.

\subsubsection{Robust linear programming}
A second and novel approach is a robust variant of linear programming (LP) for RCMDPs, extending the traditional LP framework of \cite{Altman1998}.  Note that \cite{Zheng2020} have previously used a ``robust'' version of LP in the context of UCRL, which estimates an upper confidence bound on the cost. This section proposes a similar variant of robust linear programming, with the key difference that 1) our version is discounted, non-episodic for an extension of $E^3$ whereas their version is based on an undiscounted, episodic framework for an extension of UCRL; 2) separate exploration, exploitation, and escape policies are optimised; and 3) the error term in the budget (right hand side) as opposed to the constraint-cost, although this is merely a superficial difference. A downside of this approach is that similar to \cite{Zheng2020}, this formulation does not account for uncertainty in the transition dynamics. The agent may therefore get stuck in unknown states longer than anticipated.

The approach uses the occupation measure \citep{Altman1999}, which can be interpreted as the total proportion of discounted time spent in a particular state-action pair. The occupation measure allows to formulate the asymptotic constraint-cost as a simple weighted sum of its immediate cost. For a CMDP $\hat{M}_{\mathscr{S}}$, a policy $\pi$, and a state-action pair $(s,a)$, the $T$-step estimate of the occupation measure is defined as 
\begin{equation}
f(s,a) = (1 - \gamma) \sum_{t=0}^{T} \gamma^{t} \mathbb{P}_{\pi}(s,a \vert \hat{M}_{\mathscr{S}}) \,,
\end{equation}
leading to the definition of a policy as
\begin{equation}
\pi(a\vert s) = \frac{f(s,a) }{\sum_{a' \in \mathcal{A}} f(s,a')} \,.
\end{equation}
Now the state-action value and constraint-cost functions can be expressed in terms of $f_{\pi}$, namely 
\begin{align}
\hat{V}_{\pi}(s,a, T \vert \hat{M}_{\mathscr{S}}) = f(s,a) \, \hat{r}(s,a\vert \hat{M}_{\mathscr{S}}) \nonumber
\end{align}
and
\begin{align}
\hat{C}_{\pi}(s,a, T \vert \hat{M}_{\mathscr{S}}) = f(s,a\vert \hat{M}_{\mathscr{S}}) \, \hat{c}(s,a \vert \hat{M}_{\mathscr{S}}) \,. \nonumber
\end{align}

For the exploitation and exploration policies in $E^4$, this yields the linear programming problem
\begin{equation}
\max_{f} f^T \hat{r}  \quad \text{s.t.} \\ \quad  f^{T} \hat{c} \leq l \,,
\end{equation}
where $f^T \in \mathbb{R}^{SA}$ and $\hat{r} \in \mathbb{R}^{SA}$. For a Worst-case Escape CMDP, the robust LP is not recommended due to requiring known transition dynamics but if the constraint-cost estimate is set to $c_{\text{max}}$, one may optimise the linear programming problem
\begin{equation}
\max_{f} f^T \hat{r}  \quad \text{s.t.} \\ \quad  f^{T} c_{\text{max}} \leq d' \,.
\end{equation}

Various general polynomial-time algorithms exist for linear programming, including the ellipsoid method \citep{Khachiyan1979} and  the projective method  \citep{Karamarkar1984}, which have $O(n^6L)$ and $O(n^{3.5}L)$ on $O(L)$ digit numbers. Following \cite{Karamarkar1984}, interior point methods have been further developed, providing $\epsilon$-optimal guarantees with much improved time complexity (see \cite{Potra2000} for a variety of algorithms).
\subsubsection{Dynamic programming approaches}
Dynamic programming (DP) as used in the original $E^3$ provides convergence to the optimal value and a time complexity of $O(S^2T)$. In $E^4$, DP cannot be directly applied due to the robust constrained setting. There is currently no suitable algorithm for constrained-robust DP. We propose two possible approaches but leave a full analysis for further research. The first approach starts with constrained DP  \citep{Altman1998} and then incorporates the uncertainty set while the second approach starts with robust DP \citep{Nilim2005,Iyengar2005} and incorporates constraints by reformulating the CMDP as a Lagrangian MDP \citep{Taleghan2018}.
\paragraph{Dual linear program}
The technique by \cite{Altman1998} goes as follows when applied to a CMDP $\hat{M}_{\mathscr{S}}$, which has transition dynamics $\hat{P}$, constraint-cost function $\hat{c}$, and reward function $\hat{r}$. Note that the solution to the unconstrained DP problem
\begin{align}
\phi(s) = \max_{a \in \mathcal{A}} \left[ (1-\gamma) \hat{r}(s,a) +  \gamma \sum_{s' \in \mathcal{S}} \hat{P}_{s,a}(s') \phi(s')  \right] \quad \forall s \in \mathcal{S} \,. \nonumber
\end{align}
can be rewritten as an LP of the form
\begin{align}
&\max \phi(s) \nonumber \\
& \quad \text{s.t.} \quad \phi(s) \geq (1 - \gamma) \hat{r}(s,a) + \gamma \sum_{s' \in \mathcal{S}} \hat{P}_{s,a}(s') \phi(s')  \quad \forall s \in \mathcal{S} \quad \forall a \in \mathcal{A} \,.
\end{align}
For constrained DP, the value can be defined based on the min-max of the Lagrangian
\begin{align}
\hat{V}_{\pi_{\theta}}(s,T \vert \hat{M}_{\mathscr{S}}) = \min_{\lambda \geq 0} \max_{\pi_{\theta}} L(\lambda,\pi_{\theta}) \,, \nonumber
\end{align}
where min and max can be interchanged as strong duality holds without requiring Slater's condition. Representing $L(\lambda,\pi_{\theta})$ in terms of the immediate rewards and constraint-cost results in the LP
\begin{align}
&\hat{V}_{\pi_{\theta}}(s,T \vert \hat{M}_{\mathscr{S}}) = \max_{\lambda,\phi} \phi(s)  - \lambda d \nonumber \\ 
&\quad \text{s.t.} \quad \phi(s) \geq (1 - \gamma) \left(\hat{r}(s,a) + \lambda \hat{c}(s,a)\right) + \gamma \sum_{s' \in \mathcal{S}} \hat{P}_{s,a}(s') \phi(s')  \quad \forall s \in \mathcal{S} \quad \forall a \in \mathcal{A} \,. 
\end{align}

In $E^4$, the transition dynamics are uncertain. Adding the constraints for all transition dynamics in the uncertainty set and using the estimators yields the LP
\begin{align}
&\hat{V}_{\pi_{\theta},\mathcal{P}}(s, T \vert \hat{M}_{\mathscr{S}}) = \max_{\lambda,\phi} \phi(s)  - \lambda d \nonumber \\ 
&\quad \text{s.t.} \quad \phi(s) \geq (1 - \gamma) \left(\hat{r}(s,a) + \lambda \hat{c}(s,a)\right) + \gamma \sum_{s' \in \mathcal{S}} P_{s,a}(s') \phi(s')  \quad \forall s \in \mathcal{S} \quad \forall a \in \mathcal{A} \nonumber\\
& \quad \quad \quad \forall P \in \mathcal{P} \,.
\end{align}

\paragraph{Robust dynamic programming with Lagrangian MDP}
Another proposed approach for dynamic programming in $E^4$ is to apply robust DP  \citep{Nilim2005,Iyengar2005} to a Lagrangian MDP  \citep{Taleghan2018}. Robust DP models the value as
\begin{equation}
\label{eq: RDP}
\hat{V}(s, T\vert \hat{M}_{\mathscr{S}}) = \max_{a \in \mathcal{A}} \hat{r}(s,a) + \gamma \max_{P \in \mathcal{P}_{s,a}} P^T \hat{V}(\cdot, T\vert \hat{M}_{\mathscr{S}})  \,,
\end{equation}
for worst-case transition $P \in \Delta^S$ and the values for each following state $\hat{V}(\cdot, T\vert \hat{M}_{\mathscr{S}}) \in \mathbb{R}^S$. To solve the ``inner problem'', $\max_{P \in \mathcal{P}_{s,a}} P^T \hat{V}(\cdot, T\vert \hat{M}_{\mathscr{S}})$, one uses a bisection algorithm, yielding time complexity $O(S \log(G/\epsilon_s))$, where $G$ is an upper bound to the value function and $\epsilon_s$ is the desired accuracy of the approximation. The overall problem (Eq.~\ref{eq: RDP}) gives $\epsilon$-optimal guarantees in polynomial time complexity $O(TS^2 \log(1/\epsilon)^2)$, adding only $O(\log(1/\epsilon)^2)$ time cost compared to traditional DP.

To make robust DP work for $E^4$, one can then construct a Lagrangian MDP which redefines the reward function as a linear combination of the original reward function and the constraint-cost function, such that the value is similar to the Lagrangian in Eq.~\ref{eq: Lagrange}.

A downside of using robust DP is the requirement of $(s,a)$-rectangular uncertainty sets (see Section \ref{sec: uncertainty-sets}). More general uncertainty sets can be converted into $(s,a)$-rectangular format by projection onto a larger rectangular subspace \citep{Nilim2005} but this will result in a higher upper bound on the worst-case constraint-cost.

\subsection{Exploration, Exploitation and Escape policies}
\label{sec: EEEpolicies}
The three different kinds of policies are optimised over different CMDPs introduced earlier in Section~\ref{sec: theory}. This section discusses how these policies are used, what they represent, and when they are activated.

Due to the Constrained Exploit-or-Exploit Lemma (Lemma~\ref{lem: CEE}), for any satisfiable budget $l$ and any $\epsilon \geq 0$, from a starting state $s$, either a) there exists a policy $\pi \in M_{\mathscr{S}}$ for which $V_{\pi}(s, T \vert M_{\mathscr{K}}) \geq V_{\pi^{*}}(s, T) - \epsilon$, where $\pi^{*} = \argmax_{\pi} V_{\pi}(s)  \, \text{s.t.} \, C_{\pi}(s) \leq l$ for all $s \in \mathcal{S}$; or b) there exists a policy $\pi$ in  $M_{\mathscr{K}}$ which reaches the terminal state $s_0$ in $\mathcal{S} \setminus \mathscr{K}$ in at most $T$ steps with probability $p > \epsilon /G_{\text{max}}^r(T)$. The estimated exploitation and exploration policies represent case a) and b), respectively, performing $l$-safe exploitation and $l$-safe exploration. The exploitation policy is ideally activated whenever it is known that $\hat{V}_{\hat{\pi}}(s \vert \hat{M}_{\mathscr{K}}) \geq V_{\pi^{*}}(s) - \epsilon$. In practice, this knowledge is often not available. However, one may use a strategy similar to what has been proposed for the exploitation and exploration policy in $E^3$ \citep{Kearns2002}; that is, one activates the exploration policy first and continues it as long as $p > \epsilon /G_{\text{max}}^r(T)$ remains likely, and then, as soon as $p < \epsilon /G_{\text{max}}^r(T)$ with high probability, one activates the exploitation policy which is then guaranteed to exist by Lemma~\ref{lem: CEE}. 

With the above in mind, the different policies and their corresponding induced CMDPs can now be discussed. First, the $l$-safe exploitation policy has the aim of solving a CMDP induced over the known states $M_{\mathscr{K}}$, differing from $M$ only in the sense that: 1) it terminates with $r(s_0)=0$ and $c(s_0)=0$ as soon as it reaches the terminal state $s_0$, which is when an unknown state in $\mathscr{U}$ is entered; and 2) the allowed constraint-cost budget is more limited, namely it is set to $l =  d'' - 2\epsilon$  following Lemma~\ref{lem: lEE}. Second, the $l$-safe exploration policy has the aim of solving a CMDP induced over the known states $M_{\mathscr{K}}'$. $M_{\mathscr{K}}'$  differs from $M_{\mathscr{K}}$ in the sense that it terminates with $r(s_0)=r_{\text{max}}$ as soon as it reaches the terminal state $s_0$, which is when an unknown state in $\mathscr{U}$ is entered, and that it receives $r(s)=0$ for all $s \in \mathscr{K}$. Third, the $d'$-safe escape policy has the aim of solving a CMDP induced over the unknown states $M_{\mathscr{U}}$. It differs from $M$ in the sense that: a) it terminates with $r(s_0)=0$ and $c(s_0)=0$ as soon as it reaches the terminal state $s_0$, which is when a known state in $\mathscr{K}$ is entered; and b) the allowed constraint-cost budget $d'$ is set according to Lemma~\ref{lem: EB}. Finally, to ensure constraint-satisfaction, Lemma~\ref{lem: EB} requires an additional $T$-step trajectory of safe return within the known states, a trajectory which yields a cumulative constraint-cost of at most $d_s - \epsilon$. Such a safe return policy can be similarly formulated as a CMDP (e.g. by rewarding 1 for known states and 0 for unknown states) but is often readily available from domain knowledge; for example, domains such as those in Example~\ref{ex: 2} include an action which makes the agent stay in a particular state or set of states for extended periods of time.

Before the main loop, $\epsilon > 0$ is chosen and the $\epsilon$-horizon time $T\gets \frac{1}{1 - \gamma} \ln(\frac{\max(r_{\text{max}},c_{\text{max}})}{\epsilon(1 - \gamma)})$ and safety budget $l \gets d'' - 2\epsilon$ are chosen accordingly. At the start of each iteration of the main loop, exploration and exploitation policies are optimised. Via Lemma~\ref{lem: lEE}, $l$-safe exploration or 
exploitation policies will with high probability be constraint-satisfying despite the $T$-step estimated constraint-cost in $\hat{M}_{\mathscr{K}}$ differing from the  asymptotic true constraint-cost in $M_{\mathscr{K}}$. Once an unknown state is visted, $d'$ and $T'$ are set according to Lemma~\ref{lem: EB}. As long as the different policies satisfy their respective constraint-cost 
budget within their induced CMDP, $l$ for the exploration and exploitation policies and $d'$ for the escape policy, then the overall non-stationary policy applying these policies sequentially will satisfy the constraint-cost in the full CMDP $M$.

\begin{algorithm}
\caption{E$^4$ algorithm: Explore or Exploit-Explore-Exploit-or-Escape for CMDPs.} \label{alg: E4}
\begin{minipage}[c]{0.85 \textwidth}
\begin{algorithmic}[1]
{\scriptsize
\Require starting set of known states $\mathscr{K}$, known-state budget $d''$, safe return budget $d_s$, learning rate schedules $\eta_{1}$ and $\eta_{2}$, and $\epsilon > 0$.
\State Start from random $s \in \mathscr{K}$.
\State Define $l \gets d'' - 2\epsilon$.  \Comment{$l$-safe exploration or exploitation }
\State $T = \frac{1}{1 - \gamma} \ln(\frac{\max(r_{\text{max}},c_{\text{max}})}{\epsilon(1 - \gamma)})$. \Comment{$\epsilon$-horizon time}
\While{\textbf{true}}
\LineComment{perform off-line optimisations (see Algorithm~\ref{alg: optimisation})}
\LineComment{Exploitation CMDP $\hat{M}_{\mathscr{K}}$}
\State $\hat{\pi} \gets \texttt{\sc{Offline-Optimisation}}(\hat{\pi},\hat{M}_{\mathscr{K}},\mathcal{P},l,T)$. 
\LineComment{Exploration CMDP $\hat{M}_{\mathscr{K}}'$}
\State $\hat{\pi}' \gets \texttt{\sc{Offline-Optimisation}}(\hat{\pi}',\hat{M}_{\mathscr{K}}',\mathcal{P},l,T)$. 
	\If {$\hat{V}_{\hat{\pi}}(s, T \vert \hat{M}_{\mathscr{K}}) \geq V_{\pi^{*}}(s) - \epsilon$} \Comment{Idealised ($\pi^*$ is unknown)}
	\State \Comment{in relaxed version, explore first, then exploit}
	\State $\pi \gets \hat{\pi}$. \Comment{exploit}
	\Else
	\State $\pi \gets \hat{\pi}'$. \Comment{explore}
	\EndIf
\LineComment{explore or exploit}
\For {$t=0,\dots,T-1$}
\State $a \gets \pi(s)$. \Comment{take action}
\State Sample reward $r$ and constraint-cost $c$ based on $(s,a)$.
\State $s \sim P_{s,a}^{*}$. \Comment{sample next state}
\If{ $s \in \mathscr{U}$}
\State Set budget $d'$.\Comment{Lemma~\ref{lem: EB}}
\State $T' \gets \inf \{T \in \mathbb{N}: \sum_{t=0}^{T} \gamma^{t} c_{\text{max}} \geq d'  \}$.
\State $\hat{\pi}'' \gets \texttt{\sc{Offline-Optimisation}}(\hat{\pi}'',\hat{M}_{\mathscr{U}},\mathcal{P},d',2T')$. 
\State \textbf{break}.
\EndIf
\EndFor
\LineComment{balanced wandering with escape (see Lemma~\ref{lem: WEL}, \ref{lem: SBW})}
\State $\pi \gets \hat{\pi}''$. \Comment{Worst-case Escape policy}
\State \texttt{wandering} $\gets$ \textbf{true} \Comment{start with balanced wandering}
\State Initialise path cost $C_p \gets 0$.
\For {$t=0,\dots,T'-1$}
\If{\texttt{wandering}}
\State $ a \gets \arg \min_{a' \in \mathcal{A}}  n(s,a')$. \Comment{balanced wandering}
\Else
\State $a \gets \pi(s)$. \Comment{take action from escape policy}
\EndIf
\State $n(s,a) +\!\!= 1$. \Comment{count the visitation}
\If { $n(s,a) = m_{\text{known}}$ }
\State $\mathscr{K} \gets \mathscr{K} \cup \{s\}$. \Comment{add $s$ to known states}
\EndIf
\If{ $s \in \mathscr{K}$}
\State \textbf{break}.
\EndIf
\State Sample reward $r$ and constraint-cost $c$ based on $(s,a)$.
\State $s \sim P_{s,a}^{*}$. \Comment{sample next state}
\State $C_p \gets C_p + \gamma^{t} c$. \Comment{constraint-cost of past and future}
\State $\hat{C} \gets \max_{P \in \mathcal{P}} \max_{a \in \mathcal{A}} \sum_{s \in \mathcal{S}} P_{s,a}(s') \hat{C}_{P,\pi}(s', T \vert \hat{M}_{\mathscr{U}})$.
\If { $C_p  + \gamma^{t+1} \hat{C}  \geq d' - c_{\text{max}}$ }
\texttt{wandering} $\gets$ \textbf{false}.
\EndIf
\EndFor
\State Perform safe return policy $\pi_{d_s}$ for $T$ steps within $\mathscr{K}$.
\EndWhile
}
\end{algorithmic}
\end{minipage}
\end{algorithm}

\begin{algorithm}
\caption{Offline-Optimisation of RCMDPs} \label{alg: optimisation}
\begin{minipage}[c]{0.85 \textwidth}
\begin{algorithmic}[1]
\Procedure{Offline-Optimisation}{Policy $\pi_{\theta}$, Induced CMDP $\hat{M}_{\mathscr{S}}$, uncertainty set $\mathcal{P}$, constraint-cost budget $d$, trajectory length $T$, learning rate schedules $\eta_{1}$ and $\eta_{2}$}
\State Trajectories $\xi = []$.
\For {$i=1,2,\dots,I$} \Comment{independent iterations}
\State Randomly select starting state $s \in \mathscr{S}$. 
\LineComment{Simulate worst-case trajectory}
\For { $t = 0,1,\dots,T-1$ }
\If{ $s$ is terminal}
\State \textbf{break}
\EndIf
\State Action $a \gets \pi_{\theta}(s)$.
\LineComment{deterministic cost and reward in induced CMDP}
\State $c \gets \hat{c}(s,a\vert \hat{M}_{\mathscr{S}})$. $r \gets \hat{r}(s,a\vert \hat{M}_{\mathscr{S}})$.
\State $P \gets \argmax_{P' \in \mathcal{P}} \sum_{s' \in \mathcal{S}} {P'}_{s,a}(s') \hat{C}_{\pi_{\theta},P'}(s', T \vert \hat{M}_{\mathscr{S}})$.
\State State $s' \sim P_{s,a}$. \Comment{Worst-case transition}
\State Gradient $\nabla \gets \frac{\nabla_{\theta} \pi_{\theta}(a\vert s)}{\pi_{\theta}(a\vert s)}$.
\State Trajectory $\xi \gets \xi + [s,a,c,r,s']$.
\State $s \gets s'$.
\EndFor
\State $T_{\text{stop}} \gets t$.
\LineComment{Optimise the policy}
\State Initialise $V \gets 0$.
\State Initialise $C \gets 0$.
\For {$t = T_{\text{stop}}-1,T_{\text{stop}}-2,\dots,0$ }
\State	$V \gets r_t + \gamma V $ \Comment{Objective: $\max_{\theta} V(\theta) \,\, s.t. \,\, C(\theta) \leq d$}
\State $C \gets c_t + \gamma V $ \Comment{$L(\theta) = V(\theta) - \lambda \left(C(\theta) - d \right)$}
\State $\theta \gets \theta + \eta_{1}(k) * (V - \lambda C) \nabla_t$  \Comment{$dL/d\theta$}
\State $\lambda \gets \lambda + \eta_{2}(k) * (d - C)$ \Comment{$dL/d\lambda$}
\EndFor
\EndFor
\State \textbf{return} $\pi_{\theta}$
\EndProcedure
\end{algorithmic}
\end{minipage}
\end{algorithm}

\subsection{Robust constraint-satisfiability}
\label{sec: constraint-satisfiability}
For $E^4$ to work as intended, the Worst-case Escape CMDP must have robust constraint-satisfiability. That is, when selecting the worst-case transitions from an uncertainty set $\mathcal{P}$, there must be a policy $\pi$ that satisfies $\max_{P \in \mathcal{P}} \hat{C}_{P,\pi}(s, T \vert \hat{M}_{\mathscr{U}}) \leq d'$. This depends critically on two factors. First, constraint-satisfiability depends on the CMDP of interest. The parameter of the CMDP that we analyse here is the diameter; if this parameter is limited, then the constraints are satisfiable; this holds true in simulation as well as in the real world CMDP. Second, when performing offline optimisation, robust constraint-satisfiability further depends on the uncertainty set. If the uncertainty set is too broad or does not include the true transition dynamics, then constraint-satisfiability cannot be guaranteed unless for trivial constraints (e.g. where every path goes to the known states before the budget $d'$ is exceeded). Because the transition dynamics model is not known in the unknown states, we discuss how the uncertainty set can be formed based on prior knowledge and statistical inference.
\subsubsection{The diameter of the CMDP}
\label{sec: diameter}
For constraint-satisfiability in the Worst-case Escape CMDP $M_{\mathscr{U}}$, the diameter must satisfy $D(M_{\mathscr{U}}) \leq T' = \inf \{T \in \mathbb{N}: \sum_{t=1}^{T} \gamma^{t-1} c_{\text{max}} \geq d'  \}$. In this case, the policy can escape back to $\mathscr{K}$ within at most $T'$ steps, before potentially exceeding the budget (on expectation). To ensure that at least one step of balanced wandering can be taken per attempted exploration, in line with the theory in Section \ref{sec: theory} and Algorithm~\ref{alg: E4}, the diameter must satisfy $D(M_{\mathscr{U}}) \leq T' - 1$. The robust optimisation further implies that the worst-case diameter in the uncertainty set must also be at most $D(M_{\mathscr{U}}) \leq T' - 1$. This worst-case assumption therefore can be written in terms of the uncertainty set (which may be restricted to the entries with $\mathcal{P}_{s,a}$ where $s \in \mathscr{U}$ for making the assumption weaker):
\begin{equation}
\label{eq: worst-transition}
D(\mathcal{P}) := \max_{s \neq s'} \min_{\pi}  \max_{P \in \mathcal{P}}  \mathbb{E}\left[ W(s' \vert s, \mathcal{P}, \pi) \right] \leq  T' - 1 \,.
\end{equation}
Note that this ``worst-transition diameter'' differs from the worst-case diameter of \cite{Garcelon2020} in that the worst-case is over the uncertain transition dynamics rather the policy and that the best policy is taken rather than the worst-case policy. 

The assumptions on the diameter provide a worst-case guarantee. In practice, many domains (e.g. Example~\ref{ex: 2}) have properties related to reversibility, such that the number of steps taken in unknown steps relates to the number of steps to escape back to the known states. In such cases, the diameter assumption can be significantly relaxed. In addition, more informative metrics than the diameter could potentially improve the budgetary requirements specified in Section~\ref{lem: EB}. 
\subsubsection{Uncertainty sets}
\label{sec: uncertainty-sets}
For unknown states, no samples are given at the start of the algorithm, implying that uncertainty sets constructed from their state-action visitations are too large to provide safety guarantees and to make efficient use of exploration attempts. This section discusses how to construct narrow uncertainty sets to obtain safe and efficient behaviour within the unknown states. We present existing uncertainty sets, including the $(s,a)$-rectangular uncertainty sets (see e.g., \cite{Wiesemann2013,Russel2020}), and the factor matrix representation \citep{Goyal2018}, as well as the use of expert knowledge to provide tight uncertainty sets. A gridworld example is then given as an illustration (see Example~\ref{ex: 2}).
\paragraph{(s,a)-rectangular sets}
The most well-studied class of uncertainty set is the (s,a)-rectangular set, which defines a plausible interval for each $(s,a) \in \mathcal{S} \times \mathcal{A}$. An advantage of (s,a)-rectangular set is that they provide various polynomial-time results for robust optimisation \citep{Wiesemann2013,Nilim2005}. A simple example is the set based on the L1-norm \citep{Russel2020}, which defines
\begin{equation}
\mathcal{P}_{s,a} = \{ P \in  \Delta^\mathcal{S} : \vert\vert P - \hat{P}_{s,a} \vert\vert_1 \leq \psi_{s,a}    \}\,,
\end{equation}
for all $(s,a) \in \mathcal{S} \times \mathcal{A}$. Traditionally, (s,a)-rectangular uncertainty sets are formed based on Hoeffding's inequality, defining the budget $\psi_{s,a} = \sqrt{\frac{2}{n(s,a)} \ln(\frac{SA2^S}{1 - \delta_{\psi}})}$ based on the number of visitations of the state-action pairs for failure rate $\delta_{\psi}$. This will unfortunately not work when the number of visitations is small (or even zero) in unknown states, since then the uncertainty set is prohibitively broad. However, a useful alternative for the unknown states is the Bayesian Credible Region \citep{Russel2019}, which defines the uncertainty set by finding the tightest budget with low posterior belief of failure,
\begin{equation}
\psi = \min_{\psi'} \{ \psi' \in  \mathbb{R}^+ : \mathscr{P}\left(\vert\vert P^{*}_{s,a} - \hat{P}_{s,a} \vert\vert_1 > \psi'_{s,a} \right) <  \delta_{\psi} \}\,.
\end{equation}
The posterior $\mathscr{P}$ allows injecting prior knowledge via posterior sampling, either analytically (e.g. via Dirichlet prior) or alternatively, via simulation methods (e.g. Monte Carlo Markov Chain sampling methods).
\paragraph{r-rectangular sets based on factor matrix}
More efficient than (s,a)-rectangular sets are r-rectangular sets based on a factor matrix representation, which allows to efficiently treat different state-action pairs as being correlated. An uncertainty set $\mathcal{P} \subset \mathbb{R}^{S \times A \times S}$ is generated using a factor matrix $W = \{ \mathbf{w}_1, \dots, \mathbf{w}_r\}$, a convex, compact subset of $\mathbb{R}^{r \times S}$, and a fixed set of coefficients $\mathbf{u}_{s,a}^{i}$ for all $i \in \{1,\dots,r\}$ and all $(s,a) \in \mathcal{S} \times \mathcal{A}$. Specifically, one has
\begin{equation}
\mathcal{P}_{s,a} = \sum_{i=1}^{r}  u_{sa}^{i}  \mathbf{w}_{i} \,
\end{equation}
for all $(s,a) \in \mathcal{S} \times \mathcal{A}$. 

The resulting representation is flexible; for example, if $r=SA$ and $u_{sa}^{i}$ for all $i =1,\dots,r$ one has the (s,a)-rectangular case. The factor matrix can be estimated via non-negative matrix factorisation \citep{Xu2013} although  currently this requires a relatively accurate estimate $\hat{P}_{s,a}$ \citep{Goyal2018}. In addition, factorisation has been used to provide $E^3$ with scalability to large or continuous state spaces \citep{Henaff2019}; as a result, factorisation methods seem promising for improving the scalability of $E^4$.

\paragraph{Sets based on expert knowledge: action models and local inference}
When a domain expert has a high-precision model for the transition dynamics of unknown states, much tighter uncertainty sets can be formed, which is especially useful in the early stages of the $E^4$ algorithm, when the visitations are too few to provide reliable statistics. As an illustration, we consider the case where an agent is located in a discrete state space organised along Cartesian coordinates and its available actions are moving within a local neighbourhood, as is typical in many robotics applications. The expert then formulates a set of probabilistic models, each of which is valid locally in a subset of the state space (e.g.  due to the position in the landscape).

The expert first formulates $n$ action models $g_{i}(s,a)$, $i =1,\dots,n$, which output the ``typical'' next state following action $a$ in state $s$. Then, for each $i$ a probability distribution $P_{g_{i},\tau,\mathcal{N}}$ is formed by assigning high probability $1-\tau$ to $s'=g_{i}(s,a)$ and probability $\tau/N$ to a local neighbourhood $\mathcal{N}(s'\vert s,a)$ (e.g. all states $s_e \in \mathcal{S}$ with $\vert\vert s_e-s' \vert\vert_2 \leq 2$, representing plausible action outcomes), where $N$ is the size of the neighbourhood. One may additionally iterate the definition of $P_{g,\tau,\mathcal{N}}(s,a,s')$ over different neighbourhoods and different error rates if these are uncertain. For simplicity, we illustrate this only the error rate in Example~\ref{ex: 2}, i.e. we fix the neighbourhood and define $\mathcal{P}_{g_{i}} = \{P_{g_{i},\tau,\mathcal{N}} :  \tau \leq 0.1\}$.

The resulting uncertainty set, $\{\mathcal{P}_{g_1},\dots,\mathcal{P}_{g_n}\}$, is typically small when significant domain knowledge exists. However, the size of this uncertainty set can be even further reduced by eliminating models with low probability. For this purpose, one can consider the probability of the recent path $p$ under model $g_{i}$,
$\mathbb{P}_{P_{g_{i}}}( p)$. Alternatively, the expert defines a transfer probability $\rho( g_{j} \vert s,a, g_{i})$, reflecting the probability of model $g_j$ after taking action $a$ in state $s$ for which $g_{i}$ was valid. Note that the dependencies on $s$, $a$, and/or $g_{i}$ can be dropped in the case of statistical independence.

\begin{example}
\label{ex: 2}
Let $\mathcal{S}$ be a discrete set of Cartesian $xy$-coordinates in a 10-by-10 gridworld ($S = 100$). Let $\mathcal{A}$ be the set of moves within a Von Neumann 
neighbourhood, with actions \{\texttt{north},\texttt{west},\texttt{south},\texttt{east}\} moving one step in the corresponding direction and the action \texttt{stay} remaining in the same 
cell ($A=5$). 
The optimal constrained policy is to cycle around the bounds of the gridworld, without hitting the wall. If any wall is hit, the constraint-cost is 1 and the agent remains
in the same state; otherwise the constraint-cost is 0. Hitting the wall repeatedly in quick succession, for an asymptotic constraint-cost significantly larger than $d=10$ (assuming $\gamma=0.99$), is known to risk damaging the agent. Actions fail stochastically with a probability $\tau=0.05$. Initially, we have $\mathscr{K} = \{(1,1),(2,1),\dots,(9,1)\}$ 
 and $\mathscr{U} = \mathcal{S} \setminus\mathscr{K}$ as known and unknown states, respectively. The agent is aware of being initialised in the south-west wall.
\end{example}
The expert knows there are walls but only knows opposite ends are at least 5 steps away from each other. For the interior of the gridworld, $g_1(s,a) := \texttt{co}(s) + \texttt{co}(a)$ where $\texttt{co}$ defines the coordinates 
for the state and for the action (e.g. one step north is given by $(0,1)$).  For the north, west, south, and east bounds of the gridworld, respectively, the action models $i=2,\dots,5$ make the corresponding action a null move (e.g., $g_2(s,\texttt{north}) := \texttt{co}(s)$). Each model is error-free, i.e. $\tau=0$.  
For hitting the wall, the activation of the partial uncertainty set depends on the action and the previously valid $\phi_{i}$. For example, if $a=\texttt{east}$, then $\rho(g_5\vert a) = 1$ (east bound) except when $g_3$ (west bound) was previously active.

The uncertainty set constructed from the uncertainty subsets, $\{\mathcal{P}_{g_1},\dots,\mathcal{P}_{g_{5}} \}$, where $\mathcal{P}_{g_{i}} = \{P_{g_{i},\tau,\mathcal{N}} :  \tau \leq 0.1\}$. These subsets are sufficient to model hitting the wall from the different bounds of the gridworld with the given error rate. For example, if $s=(10,1) \in \mathscr{U}$ and taking the next action \texttt{east}, the worst-case 
transition within the uncertainty set is hitting the wall with 100\% probability ($\tau=0$), a scenario considered 
by action model $g_{5}$. For other actions, the worst-case transition within the uncertainty set is hitting the wall with 10\% probability ($\tau=0.10$). These worst-cases over-estimate the constraint-cost when compared to $\tau=0.05$, which would yield only $5\%$ probability.

The current budget for unknown states is $d'=5$, allowing for 5 steps in unknown states with $c_{\text{max}}=1$. Although the diameter is significantly higher, From the first unknown state $(10,1)$, the agent will behave as follows. Given the recent trajectory, the agent has the south and east bounds as active models $\{g_4,g_5\}$ due to taking more than 5 actions east from the known south-west walls. After offline optimisation iterations on the Worst-case Escape CMDP with  uncertainty set $\{P_{g_4},P_{g_{5}} \}$, the agent performs balanced wandering for a few steps and then escapes to $(10,1)$. The agent then performs $T$ actions of $\texttt{stay}$ to ensure safe return. 

After much exploration of the gridworld, eventually the agent makes all states known save for a few in the north bound. Now the uncertainty set is narrowed down to $\mathcal{P}_{g_2} = \{P_{g_2,\tau,\mathcal{N}}: \tau \leq 0.1\}$ and moreover, there are different routes to escape. Therefore, once the agent is optimised for this uncertainty set, the agent can make much quicker progress with balanced wandering. Therefore, these remaining states will quickly become known. Once the bounds are known, a near-optimal exploitation policy can now be found for the entire CMDP.

\subsection{Practical considerations}
\label{sec: practical-considerations}
\paragraph{Setting $m_{\text{known}}$ in practice}
In general, the required number of samples for a state to be known is $m_{\text{known}} = O\left(\left(STG/\epsilon\right)^{4} \text{Var}_{\text{max}} \ln\frac{1}{\delta}\right)$. In practice, $m_{\text{known}}$ needs to be set to a fixed value, which requires uncovering constants hidden by the big O notation as well as replacing asymptotic or other unpractical assumptions with practical assumptions. First, in the Constrained Simulation Lemma, obtaining the constant $K_1$ in $\alpha = K_1 (\epsilon/(STG))^2$ requires solving exactly for the two conditions specified  in addition to the big O notation $\alpha = O(\epsilon)$, plugging in exact values of $\epsilon$, $T$, $G_{\text{max}}^r(T)$, and $G_{\text{max}}^c(T)$. $G$ will be equal to $G_{\text{max}}^r(T)$ or $G_{\text{max}}^c(T)$; since these have analogous requirements, let $G = G_{\text{max}}^r(T)$. To obtain the first requirement, first note that for most practical settings, $\alpha \leq 1$. Then, set $K_1 \leq 1/144$ to obtain
\begin{align*}
(\alpha + 2\beta)TSG_{\text{max}}^r(T) &\leq 3\sqrt{\alpha} TSG_{\text{max}}^r(T)\leq \epsilon/4\,. \nonumber			  
\end{align*}
Additionally assuming $S \geq 4$ yields the second requirement,
\begin{align}
\Delta &= \alpha / \beta = \sqrt{\alpha} \leq \epsilon/(4TG_{\text{max}}^r(T))	\,.	 \nonumber	  
\end{align}
Then, note that combining the three conditions following from Hoeffding's inequality in the Known State Lemma while plugging in $K_1$ and summing over actions, which are modelled as a constant with relatively low value, and substituting $M=\max(\{r_{\text{max}},c_{\text{max}},1\})$ appropriately instead of $\text{Var}_{\text{max}}$ (as $\epsilon \to 0$ does not hold in practice), yields a general formula for $m_{\text{known}}$:
\begin{equation}
\label{eq: mknown-practice}
m_{known} \geq \frac{1}{2} A\ln\left(\frac{2}{\delta} \right) M^2 K_1^{-2} \left( STG/\epsilon \right)^{4} \,.
\end{equation}
Note that Eq.~\ref{eq: mknown-practice} implies $E^4$ is particularly suitable for domains with limited state space. Naturally, the failure rate $\delta$ should at all times be very limited but for reducing $m_{\text{known}}$ one can dynamically change the other parameters, starting with relatively high $\epsilon$ and decreasing $\epsilon$ later on in the lifetime. This would also decrease $T$ and $G$, allowing the algorithm to reach ``satisfactory'' performance levels across a large set of states more quickly.

Given certain assumptions, one may be able to significantly reduce $m_{\text{known}}$ by using a different concentration inequality. For example, one may use Bernstein's inequality if the variance of rewards and costs are known, and relatively low compared to both $r_{\text{max}}$ or $c_{\text{max}}$. For the reward function, define the random variable $X = \vert r(s,a) - r_t \vert / m$, where $r_t$ is the observed reward at time $t$ provided $s_t=s$ and $a_t=a$. Then, note that $\sum_{i=1}^{m} X_i = \vert r(s,a) - \hat{r}(s,a) \vert$. Then, if $\text{Var}_{\text{max}} \leq \alpha$, we have
\begin{align*}
     &\mathbb{P}\left( \vert r(s,a) - \hat{r}(s,a) \vert \geq \alpha \right) \leq 2e^{-\frac{m \alpha^{2}}{2(\text{Var}_{\text{max}} + \alpha/3)}} = \delta \nonumber \\
    & m = 2 \alpha^{-2}(\text{Var}_{\text{max}} + \alpha/3) \ln(2 / \delta) \nonumber\\
\end{align*}
The requirement for the constraint-cost is completely analogous, defining $Y = \vert c(s,a) - c_t \vert / m$. For the transition dynamics one can define a Bernoulli variable $Y$ which is 1 if $(s,a)$ results in $s'$ and 0 otherwise. Then, defining the random variable $Z = \vert P^{*}_{s,a}(s') - Y \vert / m$, we have the analogous result
\begin{align*}
     m &= 2 \alpha^{-2}(\text{Var}(Z) + \alpha/3) \ln(2 / \delta) \nonumber\\ \,.
\end{align*}
Therefore, defining $\mathscr{V} = \max(\text{Var}(Z),\text{Var}_{\text{max}})$ and summing over actions, we have
\begin{equation}
\label{eq: mknown-bernstein}
m_{known} \geq 2A \alpha^{-2}(\mathscr{V} + \alpha/3) \ln(2\delta)  K_1^{-2} \left( STG/\epsilon \right)^{4} \,,
\end{equation}
which provides a decrease of a factor $\frac{1}{4} \alpha^2 M^2 / (\mathscr{V} + \alpha/3)$. As an example, plugging in $M=\max(r_{\text{max}},c_{\text{max}})=100$, $\mathscr{V}=1$, $\alpha=0.10$, a 24-fold improvement can be observed.

Finally, it is worth noting that in many settings, the reward and constraint-cost function are a deterministic, rather than stochastic, function of the state-action pair. In such cases, a single sample suffices to know the average reward and constraint-cost for a given state-action pair, and therefore the number of samples for a state to be known depends only on the result from the transition dynamics, that is, 
\begin{equation}
\label{eq: mknown-determinstic}
m_{\text{known}} =   \frac{1}{2} A\ln\left(\frac{2}{\delta}\right)K_1^{-2} \left(STG\right/\epsilon)^{4} \,.
\end{equation} 

\paragraph{From worst-case to practical assumptions}
In deriving the number of actions that need to be taken to reach a near-optimal policy for the entire CMDP, many worst-case assumptions were made. With more practical assumptions, one can get improved results for the sample complexity. First, the worst-case number of exploration attempts is $N = O(G_{max}^r(T) / \epsilon \ln(S/\delta) Sm_{\text{known}})$ can be improved by noting that one does not always need to make all states known before having the $\epsilon$-optimal solution, yielding some $S' < S$. Second, both the number of exploration attempts and $m_{\text{known}}$ can be improved by estimating $G_{max}^r(T)$ and $G_{max}^c(T)$ via empirical trajectories instead of using a generic upper bound. Third, the cost of $c_{max}$ at all time steps in the Worst-case Escape CMDP could be improved to $\hat{c}(s,a) + \alpha_u(s,a)$, where $\alpha_u(s,a)$ is the approximation error for state-action pairs that have been visited already (but are not yet known). For example, one could set $\alpha_u(s,a)$ based on the standard error $\alpha_u(s,a) = \sqrt{\frac{\text{Var}_{max}^c}{n(s,a)}}$ such that more balanced wandering steps could be taken. Note that such changes to the Worst-case Escape MDP would also allow for the application of $E^4$ to CMDPs with larger diameter. Fourth, using a separate $\alpha$ (and/or $\alpha_u$) for the reward, constraint-cost, and transition dynamics functions may account for the range and variance of these different random variables. Fifth, in practice, there will be an initial set of known states; this further reduces the number of states that need to be made known to a number $S'' < S' < S$. Sixth, many of the settings from Lemma~\ref{lem: EB} can be significantly optimised in practice: (a) the diameter requirement can be relaxed in domains such as Example~\ref{ex: 2}; (b) the current path as well as predictions of future constraint-cost may indicate when the $T$-step safe return trajectory is not needed; (c) the bounds on the budget can be improved by using the specific setting of $\gamma$ of the target CMDP instead of the worst-case setting of $\gamma=1$; and (d) the known-state and unknown-state budgets can take further information into account, such as the current number of known and unknown states. Finally, the use of adaptive sampling, which has been used in prior work on $E^3$ \citep{Domingo1999}, may also be beneficial for practical $E^4$ implementations.

\paragraph{Can one use $E^4$ for constraints on individual paths?} 
A particular trajectory $p$ starting on a state $s \in \mathcal{S}$ may yield $C(p) > d$, even if $C_{\pi}(s) \leq d$, since $C_{\pi}$ is  formulated on expectation over paths.  While $E^4$ has not been designed for constraints on individual paths, if the user desires probabilistic guarantees on the constraint-cost of paths with the same level $d$, one can form a confidence interval based on the standard deviation $S_C$, which if unknown can be upper-bounded by $S_C \leq \sum_{t=0}^{\infty} \gamma^{t} \sqrt{\text{Var}_{\text{max}}^c} = \frac{1}{1-\gamma} \sqrt{\text{Var}_{\text{max}}^c}$. Then the budget can be reformulated as $d_p \gets d - 3 S_C$ such that the budget of interest $d$ is rarely exceeded. For example, if the asymptotic constraint-cost of paths is normally distributed with mean $d_p$ and standard deviation $S_C$, then there is at most a 0.1\% probability that any such path $p$ has constraint-cost $C(p) > d$. More generally, if one has no information on the distribution, one may use concentration equalities to provide upper bounds on failure probabilities. For example, with the Chebyshev-Cantelli inequality \citep{Cantelli1928}, the same example yields
\begin{align}
\mathbb{P}( C(p) - d > 0) = \mathbb{P}( C(p) - d_p > 3 S_C) <  \frac{S_C^2}{S_C^2 + 9 S_C^2} = \frac{1}{10} \,. \nonumber
\end{align}
When it is not the case that $d - 3S_C$ is positive (and greater than $2c_{max}$), an alternative safe exploration algorithm would be required, one which is explicitly formulated to handle probability of individual failures. In $E^4$, the reasoning is that having $C(p) > d$ is dangerous but usually not catastrophic, while in the above setting $C(p) > d$ is always catastrophic. To the authors' best knowledge, handling individual failures in this manner has not been done before as this requirement is too strict, but solving this exciting challenge would often be useful in safety-critical settings. 

\section{Conclusion}
Safety is of critical concern in real world applications, and especially in RL, where an agent interacts with an initially unknown environment. Unlike game environments, failures in real world safety-critical applications will finish the lifetime of the RL agent with serious cost to the owner of the agent, and potentially, society at large. Therefore, instead of the model-free RL algorithms which are formulated with this trial-and-error setting in mind, a model-based RL approach may be more suitable for safety-critical applications. Model-based RL algorithms such as $E^3$ learn near-optimal policies with polynomial sample and time complexity, making them an attractive option for learning in the real world as the model can be used to perform offline computations without requiring too much real world trial-and-error. This paper integrates $E^3$ into a constrained Markov decision process framework to show that -- as long as the constraints are satisfiable due to a limited diameter of the CMDP -- there exists an algorithm ($E^4$) which with high probability finds a near-optimal \textit{constrained} policy within polynomial time. The $E^4$ algorithm combines the use of an explicit exploration and an explicit exploitation policy with an additional escape policy that provides a safe return for states not reliably known by the model. The algorithm is attractive for safety-critical settings, not only due to the offline computations, but also because of the non-episodic setting in which there is only a single lifetime. The simulation model allows anticipating constraint-satisfaction failures and where there is uncertainty about the true environment, distributional robustness is used to ensure the worst-case scenario does not violate the constraints. Beyond theoretical results supporting the framework, a discussion highlights offline optimisation algorithms and shows  how to formulate uncertainty sets for unknown states based on prior knowledge, empirical inference, or a combination thereof. Further practical considerations discussed include relaxing the worst-case assumptions that underlie the theory, aspects of the domain and the implementation that affect sample efficiency, and the practical interpretation of the various components of the algorithm.

A few exciting research directions emerge from this paper, including the development of scalable methods for constrained robust offline optimisation, defining uncertainty sets for unknown states, as well as the explicit use of exploration, exploitation, and escape policies within other model-based or even model-free RL algorithms. Given that the explicit separation of exploration and exploitation that characterised $E^3$ has been taken up by recent works in different learning settings, including hard-exploration \citep{Ecoffet2021} and meta-learning \citep{Liu2020}, $E^4$ may provide unique perspectives on how to solve such problems \textit{safely}.

\backmatter

\section*{Declarations}
This version of the article has been accepted for publication after peer review but is not the Version of Record and does not reflect post-acceptance improvements, or any
corrections. The Version of Record of this article is published in Machine Learning, and is available online at http://dx.doi.org/10.1007/s10994-022-06201-z.\\
\textbf{Funding} David M. Bossens was supported by the UKRI Trustworthy Autonomous Systems Hub, EP/V00784X/1. Nicholas Bishop was supported by the UK Engineering and Physical Sciences Research Council (EPSRC) Doctoral Training Partnership grant. \\
\textbf{Conflict of interest} None.\\
\textbf{Ethics approval} not applicable\\
\textbf{Consent to participate} not applicable\\
\textbf{Consent for publication} not applicable\\
\textbf{Availability of data and material} not applicable\\
\textbf{Code availability} not applicable\\
\textbf{Authors' contributions} David M. Bossens: Conceptualization, Formal analysis, Methodology, Writing - Original Draft, Project administration.
Nicholas Bishop: Conceptualization, Writing - Review \& Editing\\

\bibliography{library}

\section*{Appendix A: Constrained Simulation Lemma}
Below is the proof of the Constrained Simulation Lemma and the Known State Lemma.\\
\textbf{Constrained Simulation Lemma.} \textit{Let $T \geq \frac{1}{1 - \gamma} \ln\left(\frac{\max(r_{\text{max}},c_{\text{max}})}{\frac{\epsilon}{2}(1 - \gamma)}
\right)$,\footnote{The $\epsilon/2$- and $\epsilon$-horizon times are linearly dependent and therefore their difference vanishes in the big O notation for $\alpha$ 
\citep{Kearns2002}.}  let CMDP $\hat{M}$ be an $\alpha$-approximation of CMDP $M$ with $\alpha=O\left(\left( \epsilon / \left(STG \right)\right)^2\right)$ where $G=
\max(G_{\text{max}}^r(T),G_{\text{max}}^c(T))$, and let $\pi$ be a policy in $M$. Then for any state $s$,\\
(a) $V_{\pi}(s) - \epsilon \leq \hat{V}_{\pi}(s) \leq V_{\pi}(s) + \epsilon$;\\
 (b) $C_{\pi}(s) - \epsilon \leq \hat{C}_{\pi}(s) \leq C_{\pi}(s)+ \epsilon$.}\\
\textbf{Proof:}\\
Let $\pi$ be any policy in $M$. 
Define $\beta$-small transition as a transition for which $P^{*}_{s,a}(s') \leq \beta$. After $T$ time steps, there is a probability of at most $T*S*\beta$ that a $\beta$-small transition is crossed (since at any one time step at most $S*\beta$). The proof distinguishes between case A) the path traversed includes at least one $\beta$-small transition; and B) the path does not include any $\beta$-small transition.\\
\textbf{Case 1: at least one $\beta$-small transition} \\
If we do have $\beta$-small transitions then:
1) $T$-step walks of $\pi$ from $s$ that cross at least one $\beta$-small transition contribute at most $TS \beta G_{\text{max}}^r(T)$ (on expectation) to $V_{\pi}(s,T)$ and at most $TS \beta G_{\text{max}}^c(T)$ (on expectation) to $C_{\pi}(s,T)$.\\
2) Since $\hat{M}$ is an $\alpha$-approximation of $M$, this means that $\beta$-small transitions satisfy $P^{*}_{s,a}(s') \leq \beta + \alpha$. Therefore, $T$-step walks of $\pi$ from $s$ that cross at least one $\beta$-small transition contribute at most $TS(\alpha + \beta)G_{\text{max}}^r(T)$ (on expectation) to $\hat{V}_{\pi}(s,T)$ and at most $TS (\alpha  + \beta) G_{\text{max}}^c(T)$ (on expectation) to $\hat{C}_{\pi}(s,T)$. These also give the upper bounds to the contribution to the discrepancy of $\hat{M}$ to $M$. So, the remainder of the proof consists of choosing an appropriate $\beta$ and then solving for $\alpha$ to bound the discrepancy to the desired $\epsilon$ in the following equation:
\begin{align}
\label{eq: alpha-constraint}
(\alpha + 2\beta)TSG_{\text{max}}^r(T)\leq \epsilon/4\, \\
(\alpha + 2\beta)TSG_{\text{max}}^c(T) \leq \epsilon/4\,.
\end{align}
In the general case, we will then solve the approximation for paths with no $\beta$-small transitions and then adding the above-mentioned $\epsilon/4$ to account for the maximal contribution of such $\beta$-small transitions. 

\textbf{Case 2: no $\beta$-small transitions}\\
For $T$-step trajectories with no $\beta$-small transitions, we can convert the additive approximation
\begin{align}
P^{*}_{s,a}(s') - \alpha \leq \hat{P}(s,a,s’) \leq P^{*}_{s,a}(s') + \alpha \nonumber
\end{align}
into a multiplicative approximation
\begin{align}
(1-\Delta) P^{*}_{s,a}(s')  \leq \hat{P}_{s,a}(s') \leq (1 + \Delta) P^{*}_{s,a}(s') \,, \nonumber
\end{align}
where $\Delta = \alpha/\beta$ because $P>\beta$. To illustrate, if  $P > \beta$ then if $X \leq P + \alpha$, this means that $X \leq \alpha + \beta < (1  + \Delta)*P$.
Therefore, for any $T$-step path $p$ without $\beta$-small transitions, we have
\begin{align}
(1-\Delta)^T \mathbb{P}_{P,\pi}(p)  \leq \mathbb{P}_{\hat{P},\pi}(p) \leq (1 + \Delta)^T \mathbb{P}_{P,\pi}(p)  \,. \nonumber
\end{align}

A similar argument follows for the value function. For all state-actions $(s,a) \in \mathcal{S} \times \mathcal{A}$, we have $r(s,a) - \alpha \leq \hat{r}(s,a) \leq r(s,a) + \alpha$ and therefore
\begin{equation}
\label{eq: pathapproxV}
V(p) - T\alpha \leq \hat{V}_{\pi}(p) \leq  V(p) + T\alpha \, \nonumber
\end{equation}
holds for any path $p$. Then, since we took any arbitrary path without $\beta$-small transitions, taking the expectation over such paths will yield the same inequality, while we can further additional account for $\epsilon/4$ as the added contribution of $T$-step walks \textit{with} $\beta$-small transitions. Then converting it to a multiplicative approximation over the expected $T$-step value yields
\begin{align}
(1-\Delta)^T ( V_{\pi}(s,T) - T\alpha ) - \epsilon/4 \leq \hat{V}_{\pi}(s,T) \leq (1+\Delta)^T ( V_{\pi}(s,T) + T\alpha )  + \epsilon/4 \,. \nonumber
\end{align}
Then provide an upper bound for $\Delta$ based on a Taylor expansion:
\begin{align}
\ln\Big((1 + \Delta)^T\Big) = T \ln(1 + \Delta) = T(\Delta -\Delta^2/2 +\Delta^3/3 - \dots) \geq T\Delta/2 \nonumber
\end{align}
Requiring that 
\begin{equation}
\label{eq: requirement}
(1+\Delta)^T ( V_{\pi}(s, T) + T\alpha) + \epsilon/4 \leq V_{\pi}(s, T) + \epsilon/2
\end{equation}
we need both conditions to hold:\\
1) $(1+\Delta)^T  V_{\pi}(s,T) \leq V_{\pi}(s,T) +  \epsilon / 8$. \\
2) $(1+\Delta)^T  T\alpha \leq \epsilon/8$.\\
Solving for $\Delta$ in condition 1) yields
\begin{align}
(1+\Delta)^T ( V_{\pi}(s,T) ) &\leq V_{\pi}(s,T) +  \epsilon / 8 \nonumber \\
(1+\Delta)^T &\leq (1 +  \epsilon / (8 G_{\text{max}}^r(T))) \nonumber \\
T \Delta/2 \leq T \ln(1+\Delta) &\leq \ln\left(1 +  \epsilon / (8 G_{\text{max}}^r(T)) \right) \approx \epsilon / (8 G_{\text{max}}^r(T)) \nonumber \\
\Delta &\leq \epsilon/(4TG_{\text{max}}^r(T)) \,. \nonumber
\end{align}
Line 2 and 4 of the above imply together  that $(1+\Delta)^T$ is a constant such that $\alpha T = O(\epsilon)$.

The requirement for the constraint-cost is completely analogous. For all $(s,a) \in \mathcal{S} \times \mathcal{A}$, we have $c(s) - \alpha \leq \hat{c}(s) \leq c(s) + \alpha$ and therefore
\begin{equation}
\label{eq: pathapproxC}
C_(p) - T\alpha \leq \hat{C}_{\pi}(p) \leq  C(p) + T\alpha \,
\end{equation}
holds. Requiring that 
\begin{equation}
\label{eq: requirement-c}
(1+\Delta)^T ( C_{\pi}(s, T) + T\alpha) + \epsilon/4 \leq C_{\pi}(s, T) + \epsilon/2
\end{equation}
we need both conditions to hold:\\
1) $(1+\Delta)^T  C_{\pi}(s,T) \leq C_{\pi}(s,T) +  \epsilon / 8$. \\
2) $(1+\Delta)^T  T\alpha \leq \epsilon/8$.\\
Analogous to the value function, this yields
\begin{align}
\Delta &\leq \epsilon/(4TG_{\text{max}}^c(T)) \,,
\end{align}
and $\alpha T = O(\epsilon)$.

\textbf{Solving for $\alpha$}\\
From the requirements for the value function (see Eq.~\ref{eq: alpha-constraint}), we have $(\alpha + 2\beta)TSG_{\text{max}}^r(T)\leq \epsilon/4$.  Choosing $\beta = \sqrt{\alpha}$, we require
\begin{align}
(\alpha + 2\beta)TSG_{\text{max}}^r(T) &\leq 3\sqrt{\alpha} TSG_{\text{max}}^r(T)\leq \epsilon/4\,, \nonumber			  
\end{align}
and
\begin{align}
\Delta &= \alpha / \beta = \sqrt{\alpha} \leq \epsilon/(4TG_{\text{max}}^r(T))	\,,	 \nonumber	  
\end{align}
and $\alpha T = O(\epsilon)$. The choice of $\alpha = O\left((\epsilon/(STG_{\text{max}}^r(T)))^2\right)$ provides these results.

Similarly, we require for the constraint-cost function (see Eq.~\ref{eq: alpha-constraint}) that $(\alpha + 2\beta)TSG_{\text{max}}^c(T) \leq \epsilon/4$. Analogously, requiring 
\begin{align}
(\alpha + 2\beta)TSG_{\text{max}}^c(T) &\leq 3\sqrt{\alpha} TSG_{\text{max}}^c(T) \leq \epsilon/4\,, \nonumber		  
\end{align}
and
\begin{align}
\Delta &= \alpha / \beta = \sqrt{\alpha} \leq \epsilon/(4TG_{\text{max}}^c(T))	\,,	 \nonumber	  
\end{align}
and $\alpha T = O(\epsilon)$. The choice of $\alpha = O\left((\epsilon/(STG_{\text{max}}^c(T)))^2\right)$ provides these results.

Therefore, to meet the requirements for both the value function and the constraint-cost function, selecting $\alpha = O\left((\epsilon/(STG))^2\right)$, where $G=\max(G_{\text{max}}^r(T),G_{\text{max}}^c(T))$, yields the above requirements.

Taking the expectation over the $T$-step paths in Eq.~\ref{eq: pathapproxV} and Eq.~\ref{eq: pathapproxC}, we have
\begin{align}
V_{\pi}(s, T) - \epsilon/2 \leq \hat{V}_{\pi}(s,T) \leq V_{\pi}(s, T) + \epsilon/2 \nonumber
\end{align}
and applying the Constrained T-step estimation Lemma (Lemma~\ref{lem: CTE}) to $\hat{V}$ and $\hat{C}$, we have:
\begin{align}
\hat{V}_{\pi}(s,T) \leq \hat{V}_{\pi}(s) \leq \hat{V}_{\pi}(s, T) + \epsilon/2 \,. \nonumber
\end{align}
Similarly, we have 
\begin{align}
V_{\pi}(s,T) \leq V_{\pi}(s) \leq V_{\pi}(s, T) + \epsilon/2 \,. \nonumber
\end{align}
Therefore, we have
\begin{align}
\hat{V}_{\pi}(s) \leq \hat{V}_{\pi}(s, T) + \epsilon/2 \leq V_{\pi}(s, T) + \epsilon \leq V_{\pi}(s) + \epsilon \nonumber
\end{align}
and
\begin{align}
 V_{\pi}(s) - \epsilon \leq  V_{\pi}(s, T) - \epsilon/2 \leq \hat{V}_{\pi}(s,T) \leq \hat{V}_{\pi}(s)  \,. \nonumber
\end{align}
Analogous results follow for the constraint-cost, concluding the proof.
\qed

\textbf{Known State Lemma.} \textit{Given state $s$ has been visited 
\begin{equation*}
m = O((STG/\epsilon)^4 \text{Var}_{\text{max}} \ln(1/\delta))
\end{equation*}
times, where $G=\max(G_{\text{max}}^r(T),G_{\text{max}}^c(T))$ and $\text{Var}_{\text{max}}=\max(\text{Var}_{\text{max}}^r,\text{Var}_{\text{max}}^c)$, and from $s$ each action has been executed at least $\lfloor m / A \rfloor$ times, then with probability of at least $1 - \delta$ we have\\}
\textbf{a)} $\vert P^{*}_{s,a}(s') - \hat{P}_{s,a}(s')\vert = O((\epsilon/(STG))^2)$ for any $(s,a,s') \in \mathcal{S} \times \mathcal{A} \times \mathcal{S}$; \\
\textbf{b)} $\vert r(s,a) - \hat{r}(s,a)\vert = O((\epsilon/(STG))^2)$ for any $(s,a) \in \mathcal{S} \times \mathcal{A}$; and \\
\textbf{c)} $\vert c(s,a) - \hat{c}(s,a)\vert  = O((\epsilon/(STG))^2)$ for any $(s,a) \in \mathcal{S} \times \mathcal{A}$.\\
\textbf{Proof:}\\
\textbf{a) Transition dynamics:} 
Note that $P^{*}_{s,a}(s') \in [0, 1]$. Therefore, applying Hoeffding's inequality \citep{Hoeffding1963}, and setting $\delta$ equal to the resulting upper bound, we have
\begin{align}
&     \mathbb{P}\left(\vert P^{*}_{s,a}(s') - \hat{P}_{s,a}(s')\vert \geq t\right) \leq 2e^{-2mt^{2}} = \delta \nonumber \\
&    -mt^{2}  = \frac{1}{2}\ln\left(\frac{\delta}{2}\right) \nonumber\\
&    m = \frac{\frac{1}{2}\ln\left(\frac{2}{\delta}\right)}{t^{2}}  \,. \nonumber
\end{align}
Setting $t = \left(\epsilon/(STG)\right)^{2}$, we see that
\begin{align}
m =  \frac{\frac{1}{2}\ln\left(\frac{2}{\delta}\right)}{\left(\epsilon/(STG)\right)^{4}} = \frac{1}{2}\ln\left(\frac{2}{\delta}\right)\left(STG\right/\epsilon)^{4}  \,. \nonumber
\end{align}
Thus $O\left(\left(STG/\epsilon\right)^{4}\ln\frac{1}{\delta}\right)$ samples are needed to obtain $O\left(\left(\epsilon/(STG)\right)^{2}\right)$-~accurate estimates of each transition probability.\\
\textbf{b) Rewards:} 
Note that $G^2 \geq r_{\text{max}}^2$ and that $(STG/\epsilon)^4$ grows as $\epsilon \to 0$ so restricting $\epsilon^2 \in [0,2\text{Var}_{\text{max}}^r]$ includes the worst-case asymptotic behaviour of $(STG/\epsilon)^4 \text{Var}_{\text{max}}^r \ln(1/\delta)$. For any $\epsilon \in [0,2\text{Var}_{\text{max}}^r]$ and $t = \left(\epsilon/(STG)\right)^{2} \in [0,2\text{Var}_{\text{max}}^r/r_{\text{max}}^2]$, we have \citep{Phillips2012}:
\begin{align}
&\mathbb{P}(\vert r(s,a) - \hat{r}(s,a) \vert \geq t) \leq 2 e^{-\frac{m t^2}{4\text{Var}_{\text{max}}^r}}   = \delta  \nonumber \\
&\frac{m t^2}{4\text{Var}_{\text{max}}^r} = \ln(2/\delta) \nonumber \\
&m = 4 \ln(2/\delta) \text{Var}_{\text{max}}^r t^{-2} \,. \nonumber
\end{align}
Therefore, $O\left(\left(STG/\epsilon\right)^{4} \text{Var}_{\text{max}}^r \ln\frac{1}{\delta}\right)$ samples are needed to obtain $O\left(\left(\epsilon/(STG)\right)^{2}\right)$-~accurate estimates of the average reward for a given state.\\
\textbf{c) Constraint-costs:} Analogously, with $t = \left(\epsilon/(STG)\right)^{2} \in [0,2\text{Var}_{\text{max}}^c/c_{\text{max}}^2]$, we have:
\begin{align}
&\mathbb{P}(\vert c(s,a) - \hat{c}(s,a)\vert \geq t) \leq 2 e^{-\frac{m t^2}{4\text{Var}_{\text{max}}^c}}   = \delta \nonumber \\
&\frac{m t^2}{4\text{Var}_{\text{max}}^c} = \ln(2/\delta) \nonumber\\
&m = 4 \ln(2/\delta) \text{Var}_{\text{max}}^c t^{-2} \,. \nonumber
\end{align}
Therefore, $O\left(\left(STG/\epsilon\right)^{4} \text{Var}_{\text{max}}^c \ln\frac{1}{\delta}\right)$ samples are needed to obtain $O\left(\left(\epsilon/(STG)\right)^{2}\right)$-~accurate estimates of the average constraint-cost for a given state.\\
\textbf{Combining worst-cases: } Taking the worst case of the three conditions on $m$, and setting $\text{Var}_{\text{max}}=\max(\text{Var}_{\text{max}}^r,\text{Var}_{\text{max}}^c)$, we have
\begin{align}
m = O\left(\left(STG/\epsilon\right)^{4} \text{Var}_{\text{max}} \ln\frac{1}{\delta}\right)\,. \nonumber
\end{align}
\section*{Appendix B: Explore-or-Exploit lemma}
\textbf{Constrained Explore-or-Exploit Lemma} 
\textit{Let $M$ be any CMDP, let $\mathscr{S}$ be any subset of states $\mathscr{S} \subset \mathcal{S}$, and let $M_{\mathscr{S}}$ be the induced CMDP over $\mathscr{S}$ with a given budget $d$. For any $s \in \mathscr{S}$, for any $T$, and any $\epsilon \geq 0$, we have \textbf{either a)} there exists a policy $\pi \in M_{\mathscr{S}}$ for which $V_{\pi}(s, T \vert M_{\mathscr{S}}) \geq V_{\pi^{*}}(s, T) - \epsilon$, where $\pi^{*} = \argmax_{\pi \in \Pi_c(s,d,T)} V_{\pi}(s,T)$ is the optimal constrained $T$-step policy, and which satisfies $C_{\pi}(s, T \vert M_{\mathscr{S}}) \leq d$; \textbf{or b)} there exists a policy $\pi$ in  $M_{\mathscr{S}}$ which reaches the terminal state $s_0$ in $\mathcal{S} \setminus \mathscr{S}$ in at most $T$ steps with probability $p> \epsilon /G_{\text{max}}^r(T)$, and $C_{\pi}(s, T \vert M_{\mathscr{S}}) \leq d$.}\\
\textbf{Proof:}\\
\textbf{Case a)} $V_{\pi}(s,T \vert M_{\mathscr{S}} ) \geq V_{\pi^{*}}(s,T) - \epsilon$: \\
In this case, condition a) is satisfied.\\
\textbf{Case b)} $V_{\pi}(s,T \vert M_{\mathscr{S}} ) < V_{\pi^{*}}(s,T) - \epsilon$:  \\ 
For some set of policies $\Pi^{*}$ in $M$, we have $V_{\pi}(s,T \vert M_{\mathscr{S}}) = V_{\pi^{*}}(s,T)$ for all $\pi \in \Pi^{*}$.\footnote{$\pi$ may not be the optimal 
constrained policy; since the T-step value only approximates the asymptotic value $V(s) =  \lim_{T' \to \infty} V(s,T')$, there may be many such policies.}. Since $\pi^{*} \in 
\Pi_c(s,d,T)$, this implies via the Induced Underestimation Lemma that for a subset $\Pi_{\mathscr{S}}^{*} \subseteq \Pi^{*}$, we have additionally $C_{\pi}(s,T \vert M_{\mathscr{S}}) \leq C_{\pi^{*}}(s,T) \leq d$ for all $\pi \in \Pi_{\mathscr{S}}^{*}$. Now select $\pi \in \Pi_{\mathscr{S}}^{*}$. Decompose T-paths starting from $s$ into T-paths that only contain states in $\mathscr{S}$ ($p_k$-
type paths) and T-paths that also contain states $s \not\in \mathscr{S}$ ($p_u$-type paths):
\begin{align}
V_{\pi}(s, T) &= \sum_p \mathbb{P}_{\pi,P}[p ] V_{\pi}(p) \nonumber \\ 	
						&= \sum_{p_k} \mathbb{P}_{\pi,P}[p_k] V_{\pi}(p_k) +  \sum_{p_u} \mathbb{P}_{\pi,P}[p_u] V_{\pi}(p_u) \,. \nonumber
\end{align}
For any $p_k$-type path, we have $\mathbb{P}_{\pi,P}[p_k] = \mathbb{P}_{\pi,P^{M_{\mathscr{S}}}}[p_k]$ and $V(p_k) = V(p_k \vert M_{\mathscr{S}}) \leq V(s, T \vert M_{\mathscr{S}})$, where equalities follow from the paths all containing the known set only (since for the induced model the transition dynamics are the same and rewards outside $\mathscr{S}$ are not being considered) and the inequality follows from positive rewards and $V(s, T \vert M_{\mathscr{S}})$ considering additionally $p_u$-type paths if any. Therefore,
\begin{align}
\sum_{p_k} \mathbb{P}_{\pi,P}[p_k] V_{\pi}(p_k) &=  \sum_{p_k} \mathbb{P}_{\pi,P^{M_{\mathscr{S}}}}[p_k] V_{\pi}(p_k \vert M_{\mathscr{S}}) \nonumber \\
&\leq V_{\pi}(s, T \vert  M_{\mathscr{S}}) \nonumber \\
& < V_{\pi^{*}}(s,T) - \epsilon  \nonumber 
\end{align}
Therefore, since $V_{\pi^{*}}(s, T) = V_{\pi}(s, T) = \sum_{p_k} \mathbb{P}_{\pi,P}[p_k] V_{\pi}(p_k) +  \sum_{p_u} \mathbb{P}_{\pi,P}[p_u] V_{\pi}(p_u)$, this implies that:
\begin{align}
\sum_{p_u} \mathbb{P}_{\pi,P}[p_u] V_{\pi}(p_u) &> \epsilon \,. \nonumber
\end{align}
Since $V_{\pi}(p) \leq G_{\text{max}}^r(T)$ for any $T$-step path, it follows that
\begin{align}
\sum_{p_u} \mathbb{P}_{\pi,P}[p_u] > \epsilon / G_{\text{max}}^r(T) \nonumber
\end{align}
Therefore, condition b) is satisfied. 
\section*{Appendix C: Chernoff bound for number of attempted explorations}
For any state $s \in \mathcal{S}$, allocate at most $\frac{\delta}{S}$ probability to the state failing to become known (i.e. the state receives fewer than $m_{\text{known}}$ visits). Let $p = \epsilon/G_{\text{max}}^r(T)$,  let $N$ be the number of exploration attempts, and let $m_{\text{known}}=(1-\Delta)Np$ and $\Delta  = \frac{Np - m_{\text{known}}}{Np}$. Then, by Chernoff bounds, the probability of the number of successful exploration attempts being smaller than $m_{\text{known}}$ is bounded by
\begin{align}
\mathbb{P}[X \leq (1 - \Delta) Np ] \leq \left(\frac{e^{-\Delta}}{(1-\Delta)^{1-\Delta}}\right)^{Np} = \frac{\delta}{S}\,, \nonumber
\end{align}
which leads to a total number of exploration attempts
\begin{align}
&Np \left(\Delta + (1-\Delta)\ln(1-\Delta)  \right) = \ln\left(\frac{S}{\delta}\right) \nonumber \\   
&Np\Delta =  \ln\left(\frac{S}{\delta}\right)  \quad   N \text{ is maximal for } \Delta \to 1 \text{ and } \lim_{x \to 0+}x\ln(x) = 0  \nonumber \\
&Np - m_{\text{known}} =  \ln\left(\frac{\delta}{S} \right) \nonumber \\
&N = p^{-1} \left(\ln\left(\frac{S}{\delta}\right) + m_{\text{known}}\right) \nonumber \\
& N = O\left( p^{-1}  \ln\left(\frac{S}{\delta}\right) m_{\text{known}} \right) \,. \nonumber
\end{align}
Filling in $p = \epsilon/G_{\text{max}}^r(T)$ and repeating over all $s \in \mathcal{S}$, the desired result is obtained
\begin{align}
&N = O\left( \frac{G_{\text{max}}^r(T)}{\epsilon}  \ln\left(\frac{S}{\delta}\right) S m_{\text{known}} \right) \,. \nonumber
\end{align}
\end{document}